\documentclass{article} 
\usepackage[T1]{fontenc}
\usepackage{skill2real_preprint,times}

\usepackage{amsmath,amsfonts,bm}

\def\eqref#1{equation~\ref{#1}}

\def\1{\bm{1}}

\DeclareMathAlphabet{\mathsfit}{\encodingdefault}{\sfdefault}{m}{sl}
\SetMathAlphabet{\mathsfit}{bold}{\encodingdefault}{\sfdefault}{bx}{n}

\usepackage{hyperref}
\hypersetup{hidelinks}
\usepackage{url}
\usepackage{booktabs}
\usepackage{array}
\usepackage{float}
\usepackage{placeins}
\usepackage{needspace}
\usepackage{graphicx}
\usepackage{wrapfig}
\usepackage{algorithm}
\usepackage{algpseudocode}
\usepackage{tikz}
\usetikzlibrary{arrows.meta,positioning,fit,backgrounds}
\usepackage{enumitem}
\usepackage{listings}
\usepackage{tcolorbox}
\tcbuselibrary{listings,skins,breakable}

\newtcblisting{pvgprompt}[1]{
  listing only,
  breakable,
  enhanced,
  colback=white,
  colbacktitle=white,
  colframe=black,
  coltitle=black,
  frame hidden,
  borderline west={1pt}{0pt}{black!65},
  boxrule=0pt,
  sharp corners,
  left=9pt,right=2pt,top=5pt,bottom=5pt,
  middle=0pt,
  before skip=8pt,after skip=10pt,
  listing options={
    basicstyle=\fontfamily{lmtt}\fontsize{8.5}{10}\selectfont,
    breaklines=true,
    breakatwhitespace=true,
    columns=fullflexible,
    keepspaces=true,
    showstringspaces=false,
    upquote=true,
    aboveskip=0pt,belowskip=0pt
  },
  title={#1},
  title after break={#1 (continued)},
  fonttitle=\sffamily\bfseries\small
}

\definecolor{pvgSkillBackground}{HTML}{F8F8F8}
\definecolor{pvgSkillBorder}{HTML}{BDBDBD}
\definecolor{pvgSkillHeading}{HTML}{007C83}
\lstdefinestyle{pvgskillmarkdown}{
  basicstyle=\fontfamily{lmtt}\fontsize{8.5}{10}\selectfont,
  language={},
  morecomment=[l]{\#},
  commentstyle=\color{pvgSkillHeading},
  breaklines=true,
  breakatwhitespace=true,
  breakindent=1.5em,
  columns=fullflexible,
  keepspaces=true,
  showstringspaces=false,
  upquote=true,
  aboveskip=0pt,belowskip=0pt
}
\newtcblisting{pvgskill}[1]{
  listing only,
  breakable,
  enhanced,
  colback=pvgSkillBackground,
  colbacktitle=pvgSkillBackground,
  colframe=pvgSkillBorder,
  coltitle=pvgSkillHeading,
  boxrule=0.5pt,
  titlerule=0pt,
  sharp corners,
  boxsep=0pt,
  left=9pt,right=9pt,top=5pt,bottom=5pt,
  toptitle=5pt,bottomtitle=0pt,
  before skip=8pt,after skip=10pt,
  listing options={style=pvgskillmarkdown},
  title={\#\ #1},
  title after break={\#\ #1 (continued)},
  fonttitle=\fontfamily{lmtt}\fontsize{8.5}{10}\selectfont
}

\title{Skill2Real: Agentic Skill Learning for Zero-Shot Sim-to-Real Robot Manipulation
}

\author{
  {\small\bfseries Xincheng He\textsuperscript{1,2,4}\quad Siyu Ma\textsuperscript{1}\quad
  Chang Yu\textsuperscript{1}\quad Yunuo Chen\textsuperscript{1,3}}\\[-2pt]
  {\small\bfseries Yanjia Huang\textsuperscript{3}\quad Ying Nian Wu\textsuperscript{1}\quad
  Yin Yang\textsuperscript{4}\quad Chenfanfu Jiang\textsuperscript{1}}\\[-2pt]
  {\normalfont\scriptsize
  \textsuperscript{1}University of California, Los Angeles\; 
  \textsuperscript{2}Shanghai Jiao Tong University\;
  \textsuperscript{3}Envora\;
  \textsuperscript{4}University of Utah}
}

\preprintfinalcopy
\begin{document}

\raggedbottom

\maketitle
\fancyhead{}
\renewcommand{\headrulewidth}{0pt}
\vspace{-0.40in}

\begin{figure}[H]
  \centering
  \includegraphics[width=\linewidth,trim=0bp 0bp 0bp 0bp,clip]{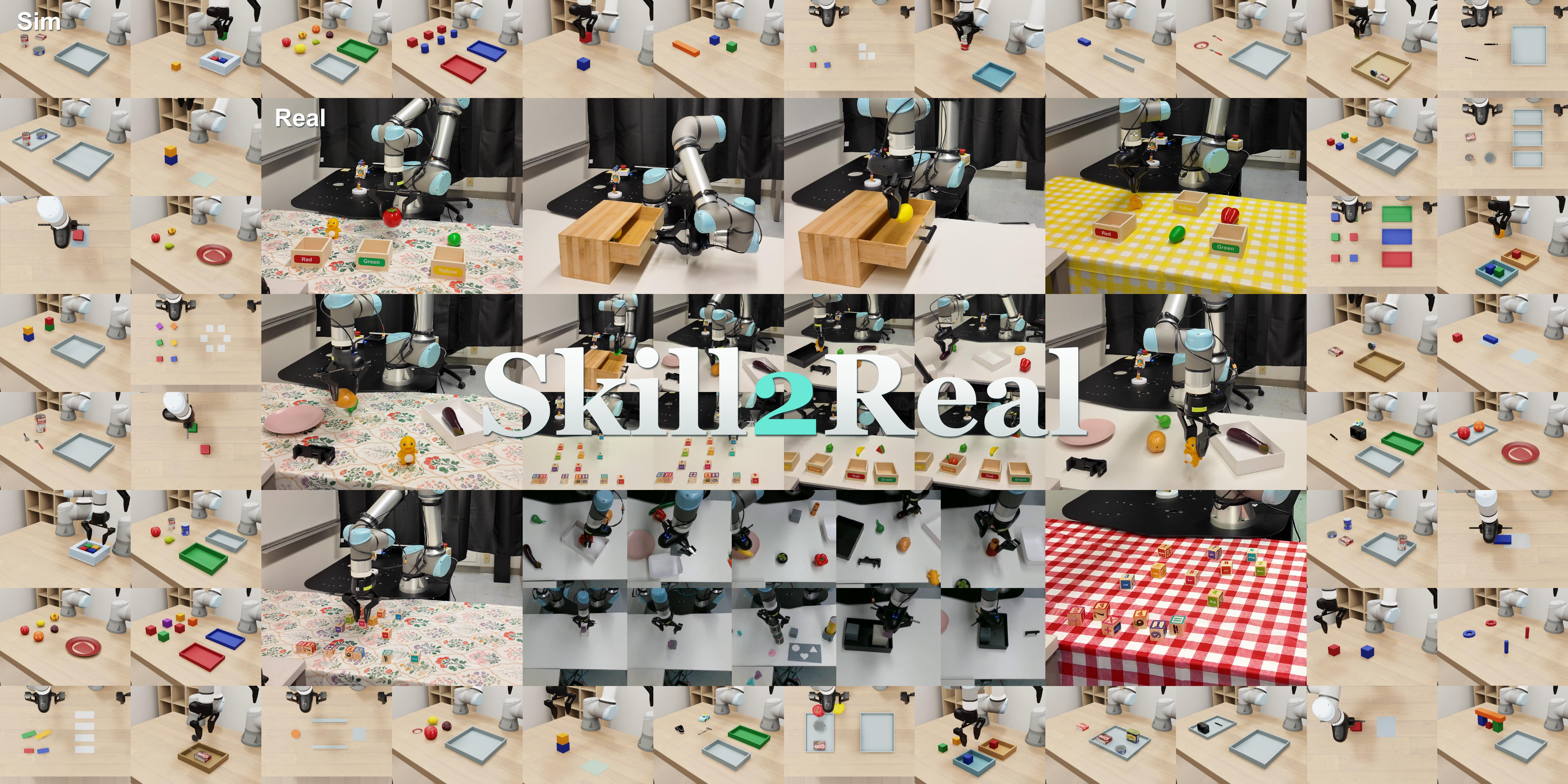}
  \vspace{-0.13in}
  \caption{\textbf{Zero-shot real-world transfer of simulation-learned skills.}
  Skill2Real learns transferable, skill-based knowledge across diverse
  simulation environments. The resulting frozen skills are evaluated on
  real-world tasks without additional learning, demonstrating robustness
  to variations in lighting, materials, and dynamics.}
  \label{fig:teaser}
\end{figure}

\begin{abstract}
Sim-to-real learning commonly adapts policies or representations to domain
variation, often directing the learning budget toward task-agnostic low-level
representations at the expense of task-level knowledge. Skill2Real builds on
cross-domain grounding from foundation vision-language models (VLMs) to
acquire validated, reusable skills in simulation through a shared
code-based policy interface. In its
asymmetric \emph{Proposer--Verifier--Governor (PVG)} loop, the Proposer acts
from public observations and API returns, the Verifier diagnoses outcomes with
privileged simulation evidence, and the Governor admits updates supported by
validation rollouts. A Cerebellum learns reusable local manipulation skills; a
Brain then composes the frozen library into long-horizon programs with planning
and recovery. At deployment, this hierarchy is grounded in real observations
through the same interface. On unseen LIBERO-Pro Long, frozen LIBERO-90
skills raise success from 2.0\% to 56.3\% with Astra and from 0.5\% to 49.0\%
with Opus 5. Robosuite success reaches 85.1--89.4\% across seven tasks. On a
real UR5e, the same agent's mean completion rises from 27.50\% to 78.75\% across
four tasks with Skill2Real, a controlled gain that, with cross-model
improvements, shows Skill2Real adds a capability orthogonal to the foundation
model by learning transferable task knowledge in simulation.
\textbf{Project page:} \href{https://skill2real.github.io}{\textcolor{blue}{skill2real.github.io}}.
\looseness=-1
\end{abstract}

\section{Introduction}
\label{section:introduction}

Zero-shot sim-to-real robot manipulation aims to convert experience acquired
in simulation into behavior that can be executed by a physical robot without
real-world task training. Simulation provides a scalable environment for
learning such behavior, but discrepancies in appearance, dynamics, contact,
and embodiment can prevent policies learned in simulation from transferring
reliably to reality~\citep{8202133,8793789}. For long-horizon manipulation,
the challenge extends beyond transferring individual actions: a robot must
reliably execute local manipulation skills, compose them into task-level
behavior~\citep{pmlr-v100-gupta20a,pmlr-v267-shi25d}, and recover when
intermediate actions fail.\looseness=-1

Existing methods leave a central bottleneck unresolved: learning reusable,
composable task knowledge from privileged simulation evidence without spending
the learning budget on task-agnostic low-level representations.  Randomization
and asymmetric actor--critic methods improve transfer by adapting neural
policies or representations~\citep{8202133,Pinto-RSS-18}.  Appearance-heavy
domain randomization can therefore spend much of the learning budget modeling
task-irrelevant variation in lighting and texture rather than acquiring
reusable manipulation knowledge.  Code-generation
methods make behavior composable, but assemble supplied primitives rather than
acquire local skills and task strategies from experience~\citep{codeaspolicies2022}.
Representative skill-library systems accumulate programs through human guidance
or real-world execution feedback~\citep{2ddda900980a441888ee44f0cdee4b2e,lu2026aspire}.
Skill2Real instead relies on foundation VLMs for cross-domain grounding and
directs simulation learning toward validated, hierarchical executable
knowledge.\looseness=-1

We introduce Skill2Real, an agentic policy framework that learns executable
robot skills through a shared code-based policy interface. The key idea is to separate the
information used to supervise skill learning from the information required to
execute learned skills: simulator privilege can evaluate and improve behavior,
while persistent skill knowledge is expressed entirely through public
observations and API semantics. The shared interface defines the transfer boundary:
programs invoke operations with common semantics, while each backend provides
its own perception, calibration, and low-level execution. This separation lets
task knowledge be reused across simulation and reality without incorporating
simulator-only state into the deployed policy.\looseness=-1

Skill2Real realizes this idea with an asymmetric
Proposer--Verifier--Governor (PVG) loop that learns a two-level skill
hierarchy. The Proposer executes programs from public observations and proposes
reusable updates; the Verifier turns privileged simulation outcomes into
publicly grounded diagnoses; and the Governor aggregates evidence across
validation rollouts to admit supported updates into memory. Stage I learns
Cerebellum skills for reusable local manipulation. Stage II freezes this
library and learns Brain skills that compose it into long-horizon task programs
with planning and recovery. At deployment, the Proposer grounds both frozen
memories in current observations through the real code-based policy
interface.\looseness=-1

Our experiments test whether frozen hierarchical knowledge generalizes across
task suites, Proposers, and the real robot. As frozen LIBERO-90
libraries~\citep{NEURIPS2023_8c3c6668} progress from no memory through the
Cerebellum to the full hierarchy, success on unseen LIBERO-Pro Long
tasks~\citep{zhou2025liberopro} rises from 2.0\% to 35.3\% to 56.3\% with
Astra and from 0.5\% to 29.3\% to 49.0\% with Opus 5, showing that both skill
levels transfer across Proposers beyond the source tasks. Across seven single-arm and bimanual
Robosuite tasks~\citep{zhu2020robosuite}, independently learned libraries raise
mean success from 16.6\% to 85.1\% with Sol and from 19.4\% to 89.4\% with
Opus 5, demonstrating consistent gains across Proposers. On the real UR5e, the
full Brain--Cerebellum hierarchy reaches 78.75\% mean task completion versus
56.25\% with local skills alone, confirming the value of task-level composition
for long-horizon behavior.\looseness=-1

Our main contributions are:

\begin{itemize}[leftmargin=*]

    \item We introduce Skill2Real, an agentic policy framework that learns
    executable robot programs in simulation and transfers frozen skill
    knowledge through a shared code-based policy interface using only public observations at
    deployment, without real-world task-policy adaptation.

    \item We develop an asymmetric Proposer--Verifier--Governor scheme: the
    Proposer acts on public observations, the Verifier translates privileged
    execution evidence into public feedback, and the Governor admits updates
    supported by validation rollouts.

    \item We develop a hierarchical policy learning strategy in which the
    Cerebellum learns reusable local manipulation skills and the Brain learns
    long-horizon compositions conditioned on the frozen Cerebellum library,
    enabling task-level planning and recovery.
\end{itemize}

\section{Related Work}
\label{section:related_work}

\paragraph{Sim-to-Real Robot Learning.} Classical sim-to-real methods
primarily bridge visual, dynamics, and execution gaps by learning policies or
representations that are invariant to domain shift. Domain
randomization~\citep{8202133} and dynamics
randomization~\citep{Sim2Real2018} expose the learner to controlled variation,
while privileged critics~\citep{Pinto-RSS-18}, real-world
adaptation~\citep{8793789}, and active
randomization~\citep{pmlr-v100-mehta20a} provide additional training signals
or target-domain experience when the gap is difficult to model. Foundation
vision-language-action models instead bring broadly pretrained
representations that can reason across heterogeneous observations, tasks, and
embodiments: RT-2~\citep{pmlr-v229-zitkovich23a} and
Open X~\citep{OpenX24} scale such learning, while
Octo~\citep{Ghosh-RSS-24} and OpenVLA~\citep{pmlr-v270-kim25c} make generalist
policies more accessible. Building on the cross-domain reasoning capability
of foundation VLMs, Skill2Real learns transferable task-level knowledge in
the form of validated, executable skills and task programs, without learning
a new representation from scratch. The frozen skills are deployed through a
shared code-based policy interface:
privileged state and evaluation are used only during simulation training,
while the real backend supplies perception, calibration, and low-level
execution.\looseness=-1

\paragraph{Language Agents and Code Policies.} Language-conditioned policies provide a natural interface for composing robot
behavior. Code as Policies~\citep{codeaspolicies2022} and
ProgPrompt~\citep{10161317} generate executable programs, while
SayCan~\citep{pmlr-v205-ichter23a} grounds language plans in physically feasible
affordances. Recent agentic systems expand beyond fixed skill sets: Agentic
Skill Discovery~\citep{zhao2024agentic} bootstraps skills and success criteria,
CaP-X~\citep{fu2026capx} studies feedback-driven coding-agent improvement,
and ASPIRE~\citep{lu2026aspire} executes, diagnoses, repairs, and validates
robot programs. Playful Agentic Robot Learning~\citep{zhang2026playful} and
ENPIRE~\citep{xiao2026enpireagenticrobotpolicy} further investigate autonomous
exploration and real-world policy evolution. Zetta~\citep{ding2026zetta}
evolves code-based runtime critics and recovery skills around frozen base
policies, combining action-level monitoring with validation-gated harness
updates. It also demonstrates zero-shot transfer of learned recovery skills.
Skill2Real is a multi-agent decision-making system in which the Proposer
generates executable robot programs, the Verifier diagnoses their outcomes
using privileged simulation evidence, and the Governor decides which validated
updates enter persistent skill memory. Together, the agents learn a two-level
hierarchy of local skills and task programs that is deployed as frozen,
publicly grounded knowledge through the shared interface.\looseness=-1

\paragraph{Robot Skill Learning and Hierarchical Policies.} Hierarchical reinforcement learning separates short-horizon control from
long-horizon decisions. Options~\citep{sutton1999between},
MAXQ~\citep{dietterich2000hierarchical}, and
HIRO~\citep{NEURIPS2018_e6384711}
establish temporal abstraction through subpolicies, value decompositions, and
subgoal-conditioned control. Relay Policy Learning~\citep{pmlr-v100-gupta20a}
and SPiRL~\citep{pmlr-v155-pertsch21a} show that reusable stages or skill
priors can improve long-horizon manipulation. More recent systems build
semantic or persistent skill libraries, including Lifelong Robot Library
Learning~\citep{2ddda900980a441888ee44f0cdee4b2e},
AtomSkill~\citep{zhu2025learning}, and Hi Robot~\citep{pmlr-v267-shi25d};
MemoryVLA~\citep{shi2025memoryvla}, MemoryVAM~\citep{jiang2026memoryvam},
RoboHarness~\citep{huang2026roboharness}, RoboHiMan~\citep{chen2025robohiman},
RoboPARA~\citep{duan2026roboparadualarmrobotplanning}, and
RigPI~\citep{he2026rigpidynamicparameteridentification} study memory, policy routing,
hierarchical evaluation, task planning, or dynamics identification for manipulation. Their
transfer units are typically policies, latent skills, fixed stages, or
episodic records. Skill2Real instead learns a two-level hierarchy of reusable
skills. The Cerebellum acquires and freezes validated local manipulation
skills, while the Brain learns task-level skills that compositionally
recombine the local library into long-horizon programs. Reuse and composition
at both levels support generalization across tasks, scenes, and deployment
domains through the shared interface.\looseness=-1

\section{Method}
\label{section:method}

Skill2Real transfers a frozen hierarchy of executable Cerebellum and Brain
skills learned in simulation, rather than model parameters. This transfer
builds on the premise that foundation VLMs provide representations capable of
zero-shot generalization from simulated to real-world visual domains, while
the shared code-based policy interface gives learned skills a common
executable form in both. The Cerebellum
learns reusable local manipulation skills, while the Brain learns task-level
skills that compose them into long-horizon programs.\looseness=-1

\begin{figure*}[t]
  \centering
  \setlength{\abovecaptionskip}{0pt}
  \includegraphics[width=\textwidth,keepaspectratio,trim=19.5bp 140bp 19.5bp 36bp,clip]{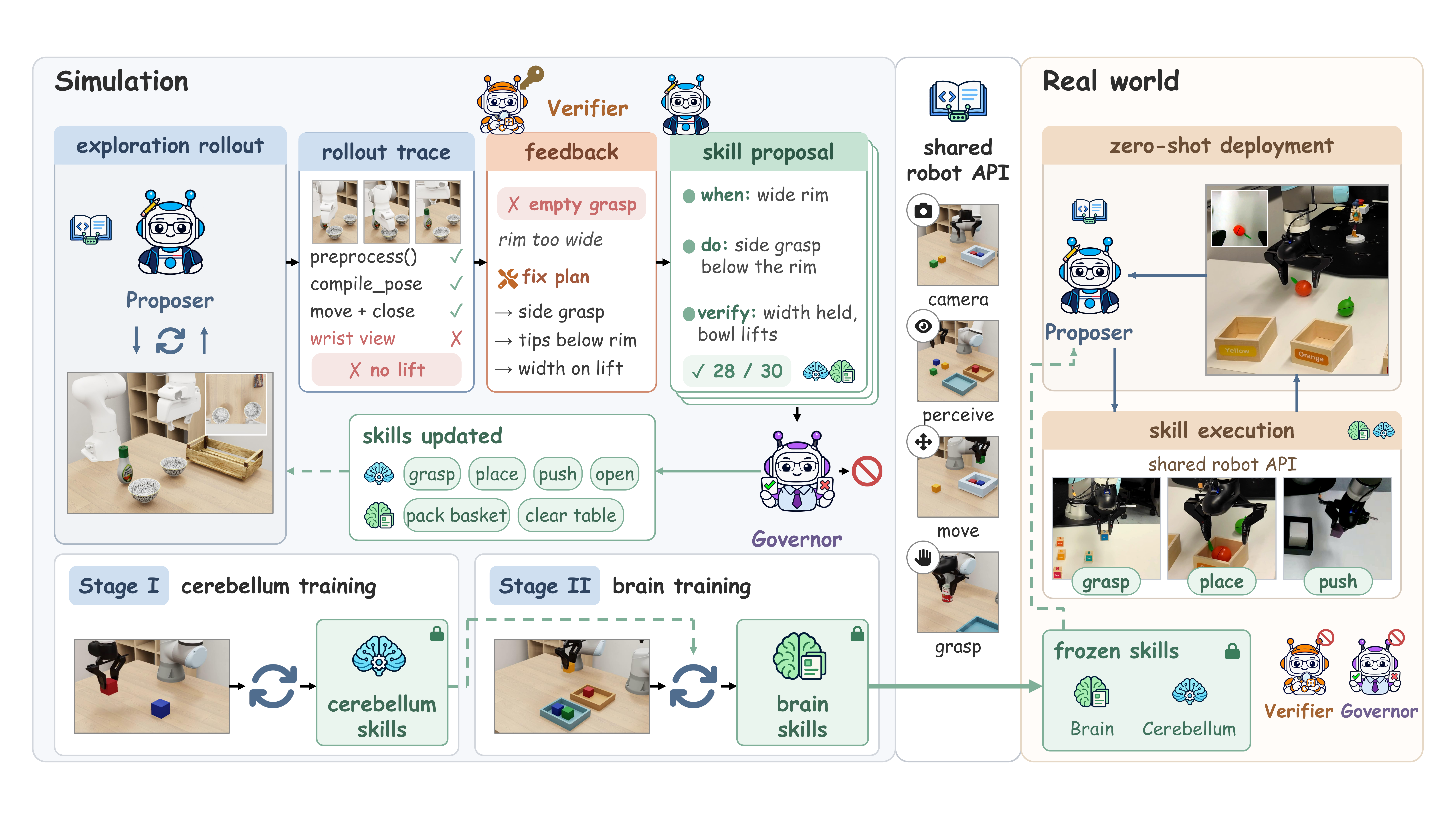}
  \vspace{-6pt}
  \caption{\textbf{Skill2Real pipeline.} All in-context skill learning occurs
    in simulated scenes. Stage I learns reusable Cerebellum skills for local
    manipulation; Stage II freezes the Cerebellum and learns Brain skills that
    compositionally organize the local library into task programs. In both
    stages, the Proposer generates programs from public observations, the
    Verifier converts privileged execution evidence into publicly grounded
    feedback, and the Governor admits only updates supported by validation
    rollouts. }
  \label{fig:skill2real_pipeline}
\end{figure*}

\subsection{Shared Code-Based Policy Interface}
\label{subsection:agentic_policy}

The central intuition is to make learned robot knowledge executable and
reusable across domains by representing it as programs over a public interface
with common semantics.

Let $\mathcal{E}^{\mathrm{sim}}$ and $\mathcal{E}^{\mathrm{real}}$ denote a
simulated and a real robot environment.  Both expose an API contract
$\mathcal{A}=\{a_k\}_{k=1}^{K}$, where $K$ is the number of documented
operations and $a_k$ is the $k$th operation.  The contract provides public
observations and robot operations with common semantics.  Its implementation
is environment-specific, but the policy reasons only through the shared
interface.  This API therefore serves as the transfer boundary between
high-level task knowledge and embodiment-specific execution.  Superscripts
$\mathrm{sim}$ and $\mathrm{real}$ identify the simulation and real-world
domains, respectively.\looseness=-1

At decision stage $t$, let $g$ be the language goal, $o_{\leq t}$ the history
of public observations and API returns through stage $t$, and
$D_{\mathcal{A}}$ the documentation for $\mathcal A$.  We use
$S_{\mathrm{cer}}$ and $S_{\mathrm{brain}}$ for the active cerebellum and
brain memories.  The Proposer, instantiated by a foundation model $M$, samples
an executable program $c_t$ according to
\begin{equation}
  c_t \sim \pi_M\!\left(c \mid g,o_{\leq t},D_{\mathcal{A}},
  S_{\mathrm{cer}},S_{\mathrm{brain}}\right).
  \label{eq:agentic_policy}
\end{equation}
Here, $\pi_M$ is the conditional distribution over candidate programs $c$
induced by $M$, and $\sim$ denotes sampling from that distribution.
Executing $c_t$ produces documented API outcomes and subsequent public
observations, which are appended to the interaction history. Consequently,
the policy is grounded in the same information channel during
simulation training and real deployment.\looseness=-1

This interface is the concrete transfer boundary: the policy transfers
executable behavior and task knowledge, while each backend implements
embodiment-specific perception, calibration, and low-level control.\looseness=-1

\newlength{\experimenttilewidth}
\newlength{\experimentleftwidth}
\newlength{\experimentrightwidth}

\begin{figure}[H]
  \centering
  \setlength{\experimenttilewidth}{\dimexpr(\linewidth-12pt)/5\relax}
  \setlength{\experimentleftwidth}{\dimexpr2\experimenttilewidth+3pt\relax}
  \setlength{\experimentrightwidth}{\dimexpr3\experimenttilewidth+6pt\relax}
  \begin{tabular}{@{}c@{\hspace{3pt}}c@{\hspace{3pt}}c@{\hspace{3pt}}c@{\hspace{3pt}}c@{}}
    \includegraphics[width=\experimenttilewidth]{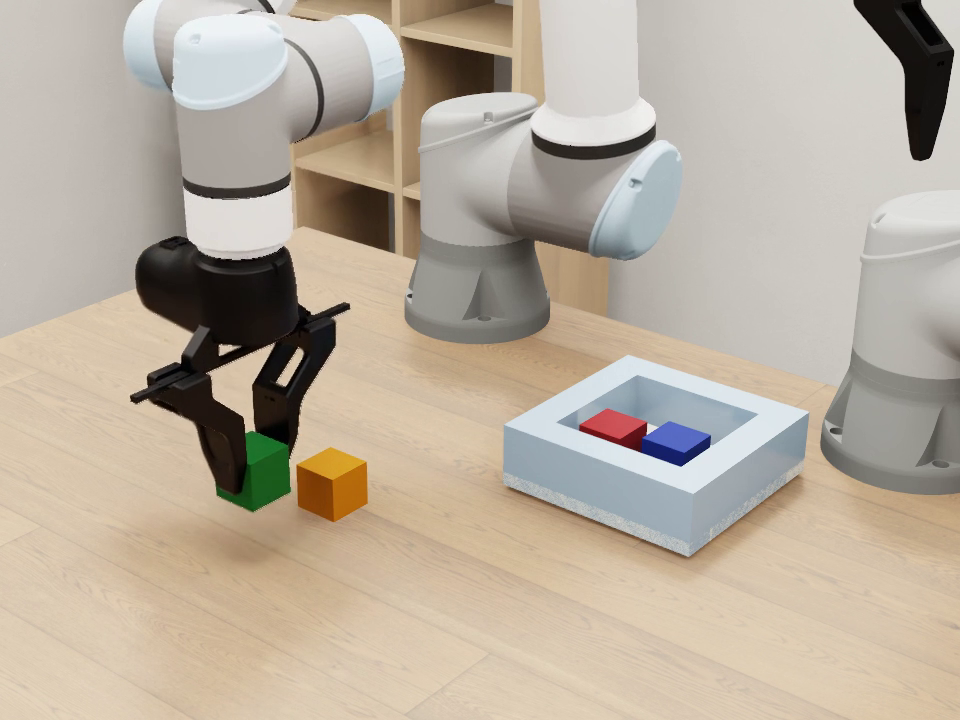} & \includegraphics[width=\experimenttilewidth]{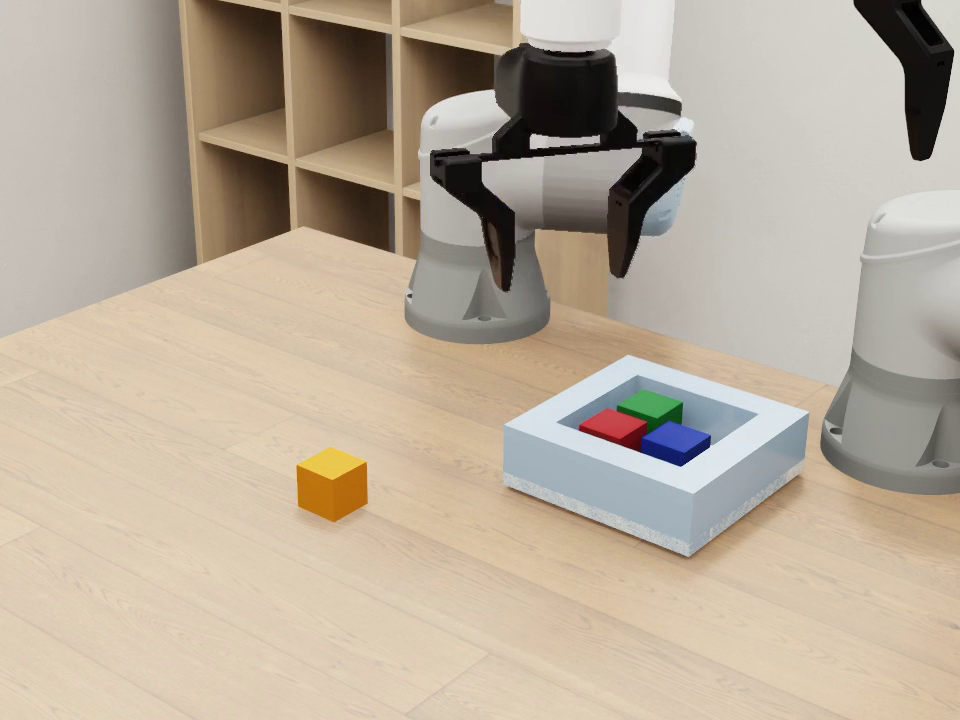}
      & \includegraphics[width=\experimenttilewidth]{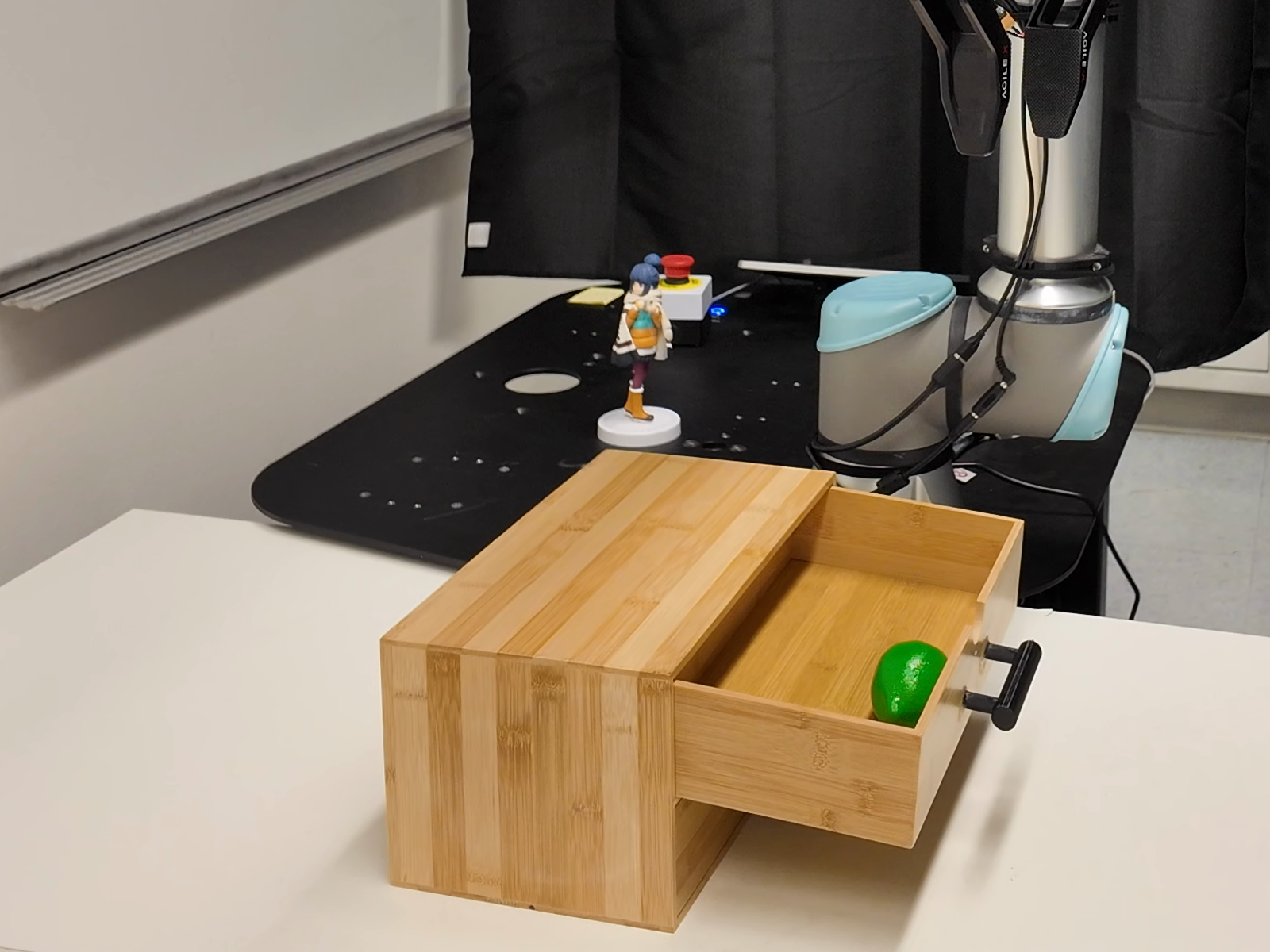} & \includegraphics[width=\experimenttilewidth]{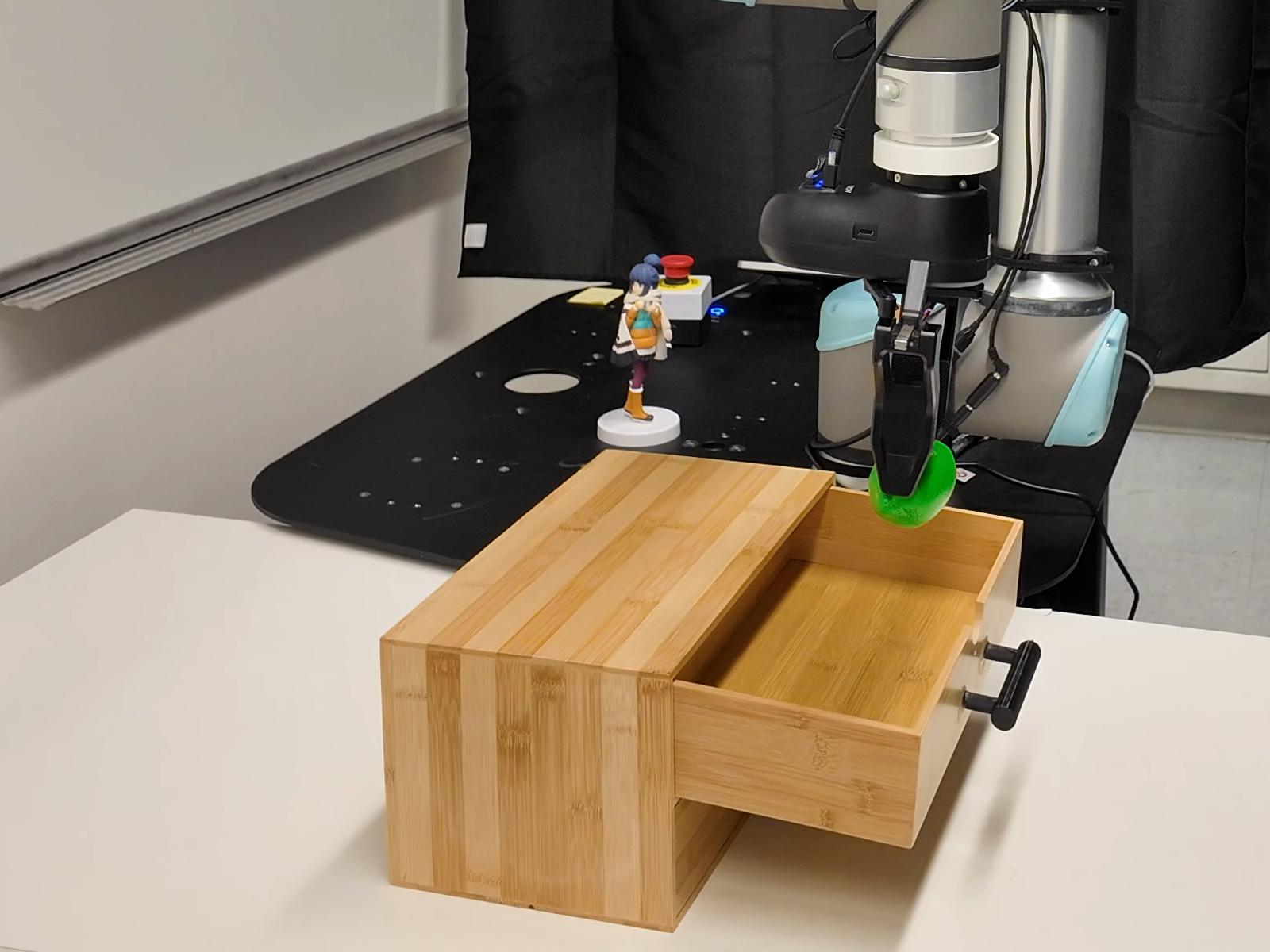}
      & \includegraphics[width=\experimenttilewidth]{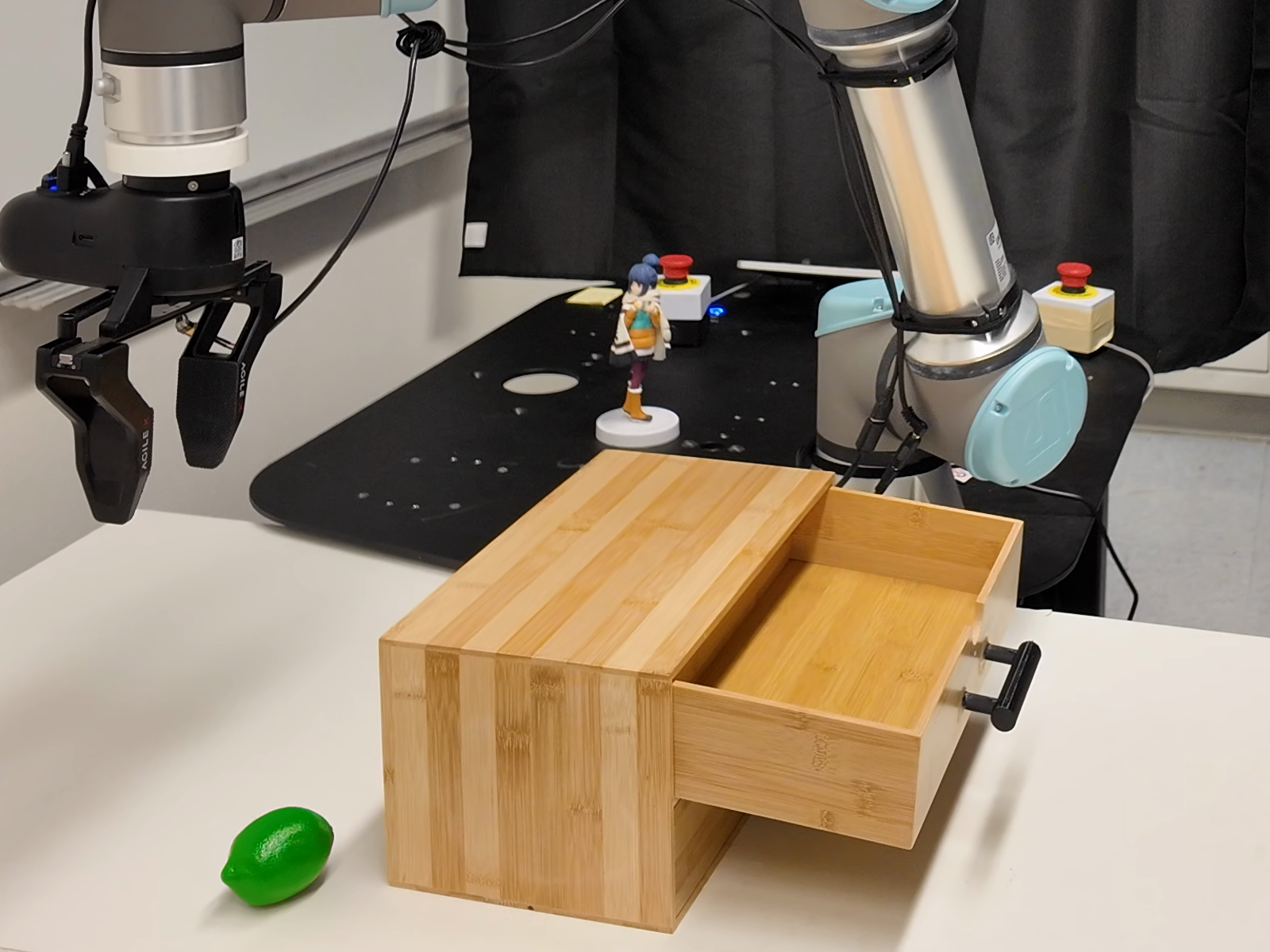} \\[1pt]
    Grasping & Placement & In drawer & Lift out & Released \\[3pt]
    \includegraphics[width=\experimenttilewidth]{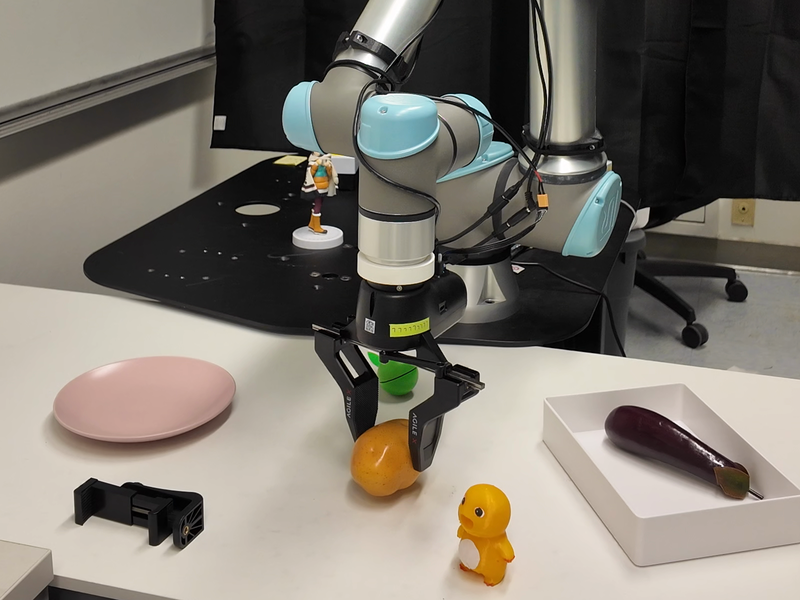} & \includegraphics[width=\experimenttilewidth]{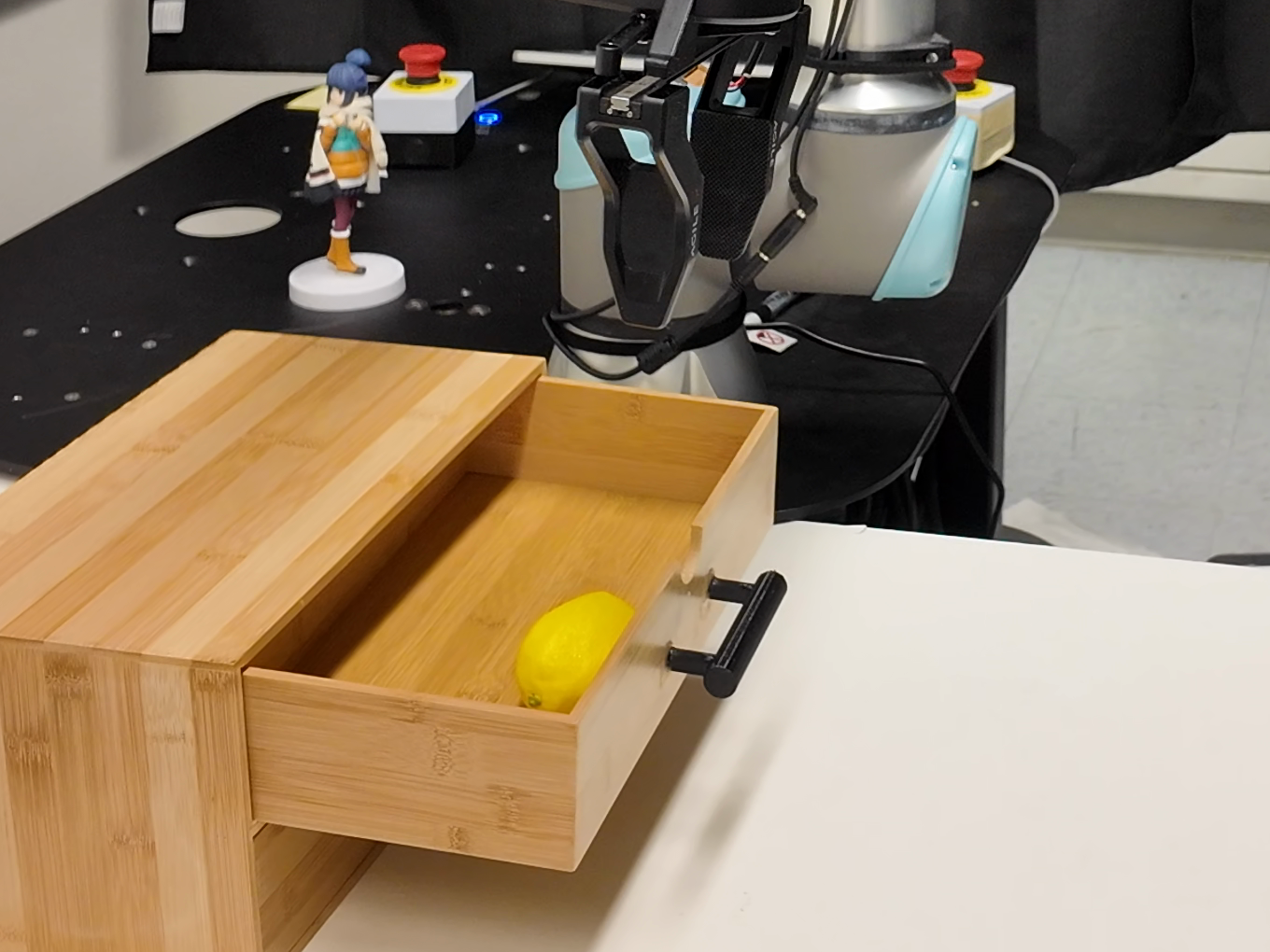}
      & \includegraphics[width=\experimenttilewidth]{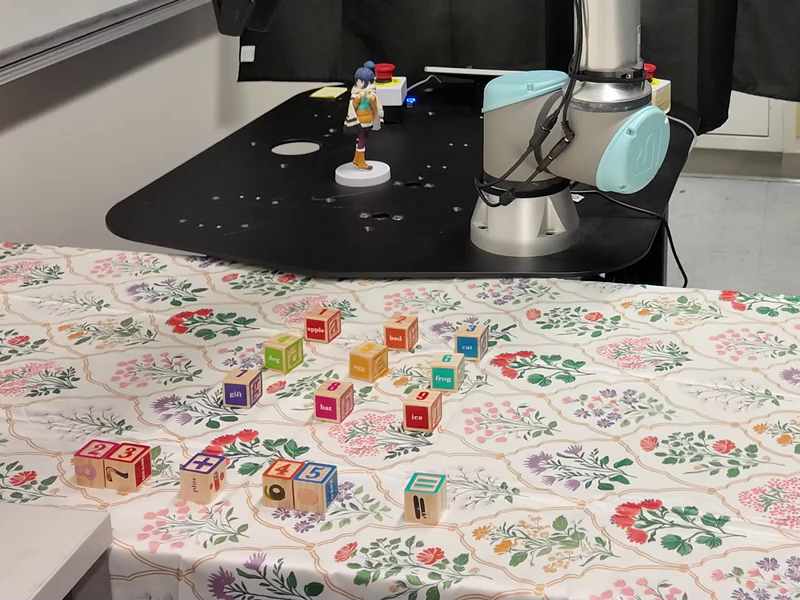} & \includegraphics[width=\experimenttilewidth]{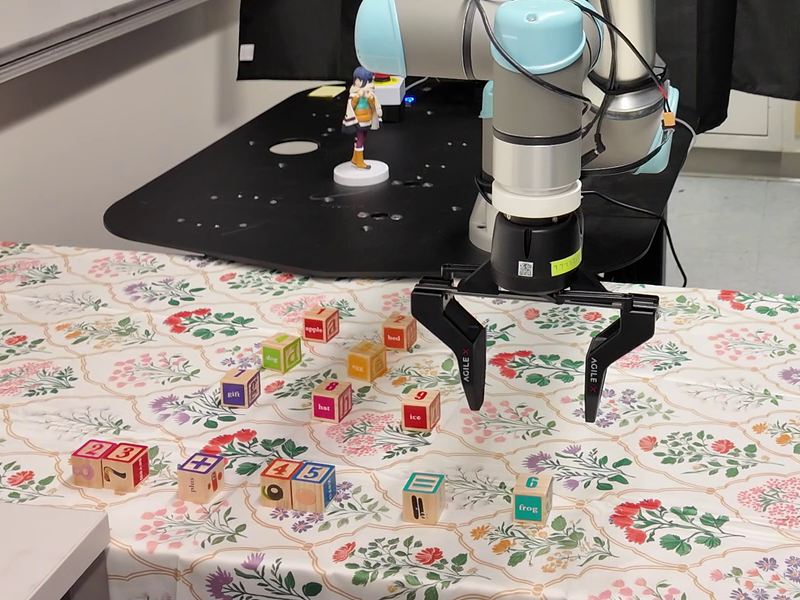}
      & \includegraphics[width=\experimenttilewidth]{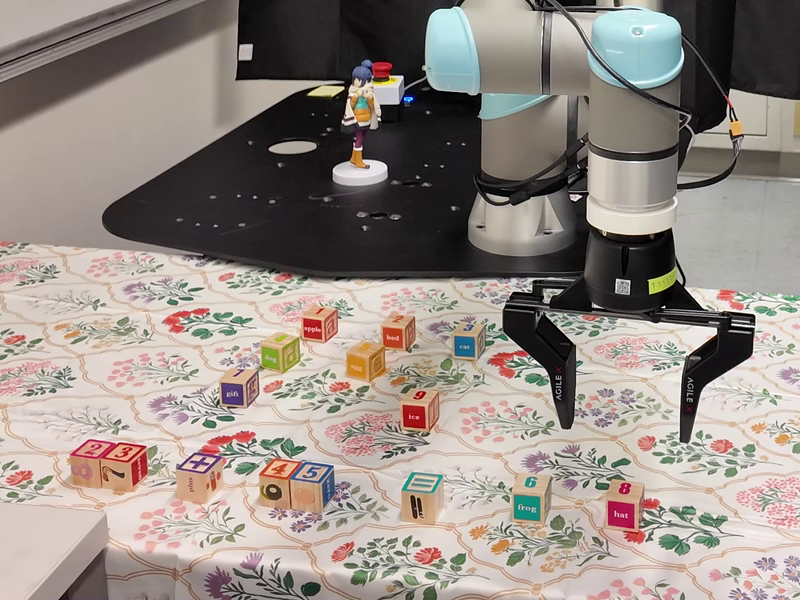} \\[1pt]
    Grasping & Placement & Initial & First placement & Final placement
  \end{tabular}\par
  \noindent\makebox[\linewidth][l]{%
  \begin{minipage}[t]{\dimexpr\experimentleftwidth-4pt\relax}
    \caption{\textbf{Cerebellum skills.} \mbox{\textit{(top)}} Pick-and-place
    in our tabletop simulator. \mbox{\textit{(bottom)}} Real-world pick-and-place.}
    \label{fig:sim_real_skills}
  \end{minipage}\hspace{7pt}%
  \begin{minipage}[t]{\experimentrightwidth}
    \caption{\textbf{Real deployment.} \textit{(top)} Complete citrus retrieval
    from the open drawer through release.
    \textit{(bottom)} R3 equation assembly ($23+45=68$).
    Each row follows one recorded execution.}
    \label{fig:real_execution}
  \end{minipage}%
  }
\end{figure}

\subsection{Multi-Agent Learning Framework}
\label{subsection:multi_agent_framework}

The multi-agent design separates roles by their information access and
learning responsibilities.  To remain deployable in the real world, the
Proposer generates programs and skill updates using only information grounded
in public observations and API outcomes available in both domains.  The
Verifier may instead inspect
privileged simulation evidence to provide stronger rollout-level supervision,
while the Governor aggregates and organizes multiple Verifier judgments
across repeated rollouts, admitting only updates with consistent empirical
support.  These three large language model (LLM) agents form the
Proposer--Verifier--Governor (PVG) loop used at both hierarchy levels.  A
deterministic execution evaluator and skill store support learning but are not
additional agents.  Deployment retains only the Proposer, shared interface,
and frozen skill memories.\looseness=-1

For a rollout trace $\tau$, let $T$ be its final decision stage and
$x_{0:T}^{\mathrm{priv}}$ the corresponding simulator-only state trajectory.
Let $\mathcal E_{\mathrm{exec}}$ denote the deterministic execution-evaluator
mapping, and let $\mathcal F_{\mathrm{pub}}$ be the set of feedback messages
expressible only through public observations and API semantics.  The
asymmetric flow is
\begin{equation}
  e=\mathcal E_{\mathrm{exec}}(\tau,x_{0:T}^{\mathrm{priv}}), \qquad
  f=\operatorname{Verifier}(\tau,e), \qquad
  f\in\mathcal F_{\mathrm{pub}}.
  \label{eq:asymmetric_feedback}
\end{equation}
Here, $e$ is the evaluator's summary of the realized physical outcome,
$\operatorname{Verifier}(\tau,e)$ denotes the Verifier's feedback mapping, and
$f$ is the resulting feedback message.  The superscripts $\mathrm{priv}$ and
subscript $\mathrm{pub}$ denote privileged and public information,
respectively. The Verifier is constrained to express \(f\) only through public observations and API semantics, so simulator-only quantities are excluded from the feedback supplied to the Proposer. Privileged
information thus supervises learning without being required at deployment.\looseness=-1

\subsection{Hierarchical Skill Training}
\label{subsection:hierarchical_training}

Long-horizon manipulation benefits from separating knowledge by abstraction:
local interaction skills recur across tasks, while task-level knowledge
specifies how to compose them for a goal.  The cerebellum memory $S_{\mathrm{cer}}$ contains primitive-level manipulation skills for reliable local interaction.  Each skill combines
natural-language guidance, an API-level code template, and a lightweight
structured state description.  The brain memory $S_{\mathrm{brain}}$ instead
contains task-level skills that compose cerebellum skills into long-horizon
behavior.  Thus, the cerebellum captures how to realize local effects, whereas
the brain captures how those effects should be organized to solve a task.\looseness=-1

The two levels share the same PVG training mechanism but differ in representation granularity and training context. Their hierarchical dependency is introduced by sequential training: Stage II learns task-level composition conditioned on the frozen Stage-I cerebellum library. Let $\mathcal Q_{\mathrm{cer}}$ and
$\mathcal Q_{\mathrm{brain}}$ denote the primitive-level and task-level
training-task distributions.  We write
$\operatorname{TrainStage}(\mathcal Q,C)$ for the shared PVG procedure that
learns a new memory from tasks sampled from $\mathcal Q$ while receiving fixed
cross-level context $C$; $C=\emptyset$ means that no such context is supplied.
Stage I runs this procedure from an empty memory on
$\mathcal Q_{\mathrm{cer}}$ and freezes the resulting cerebellum library.
Stage II starts a new empty brain memory, runs the same procedure on
$\mathcal Q_{\mathrm{brain}}$, and receives the frozen cerebellum library only
as context.  Brain training therefore learns composition without modifying the
local skills on which it depends.  This dependency is
\begin{equation}
  S_{\mathrm{cer}}^*=\operatorname{TrainStage}
  (\mathcal Q_{\mathrm{cer}},\emptyset), \qquad
  S_{\mathrm{brain}}^*=\operatorname{TrainStage}
  (\mathcal Q_{\mathrm{brain}},S_{\mathrm{cer}}^*).
  \label{eq:hierarchical_training}
\end{equation}
The superscript $*$ marks the validated memory returned and frozen at the end
of a training stage.\looseness=-1

To distinguish the evidence used to form a proposal from the evidence used to
admit it, we denote by $e_{\delta}$ the validation evidence associated with a
candidate update $\delta$.  After $\delta$ is synthesized, the updated memory
is evaluated on subsequent rollouts, and $e_{\delta}$ aggregates their
execution outcomes, such as task completion, failure, and regression relative
to the pre-update memory.  The Governor decides whether to accept or reject
the proposal based on $(\delta,e_{\delta})$, so it is admitted only when its observed
validation behavior supports the proposed improvement.  As with $e$, this
evidence is used by the learning-time Governor and is not exposed as
simulator-state input to the deployed policy.\looseness=-1

The complete PVG training procedure is provided in
Algorithm~\ref{alg:skill2real} (Appendix~\ref{appendix:pvg_algorithm}).

\subsection{Zero-Shot Sim-to-Real Deployment}
\label{subsection:deployment}

After training, Skill2Real transfers the learned skill memories through the
shared interface between simulation and reality.  The Proposer conditions on
real observations together with $S_{\mathrm{cer}}^*$ and
$S_{\mathrm{brain}}^*$.  Because both memories are expressed through public
observations and API semantics, the same hierarchy can be grounded by the real
robot.\looseness=-1

\Needspace{7\baselineskip}
\section{Experimental Setup}
\label{section:experimental_setup}

\subsection{Simulation Setup}
\label{subsection:sim_setup}

\begin{wrapfigure}[18]{R}{0.43\textwidth}
  \vspace{-10pt}
  \centering
  \pgfmathsetlengthmacro{\setupTileWidth}{(\linewidth-2pt)/2}
  \pgfmathsetlengthmacro{\setupTileHeight}{15*\setupTileWidth/17}
  \begin{tikzpicture}[inner sep=0pt,outer sep=0pt]
    \node[anchor=north west] (libero) at (0,0)
      {\includegraphics[trim=60bp 0bp 60bp 0bp,clip,
        width=\setupTileWidth,height=\setupTileHeight]{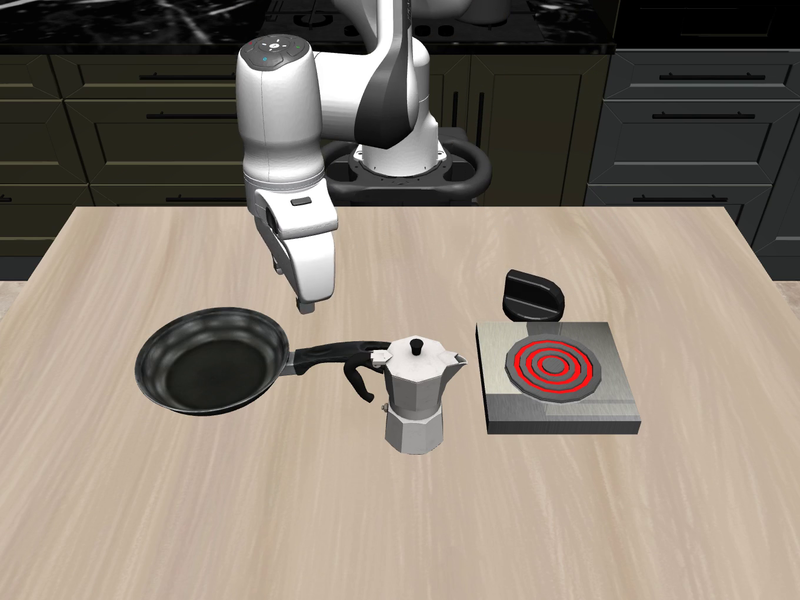}};
    \node[anchor=north west] (robosuite) at ([xshift=2pt]libero.north east)
      {\includegraphics[trim=60bp 0bp 60bp 0bp,clip,
        width=\setupTileWidth,height=\setupTileHeight]{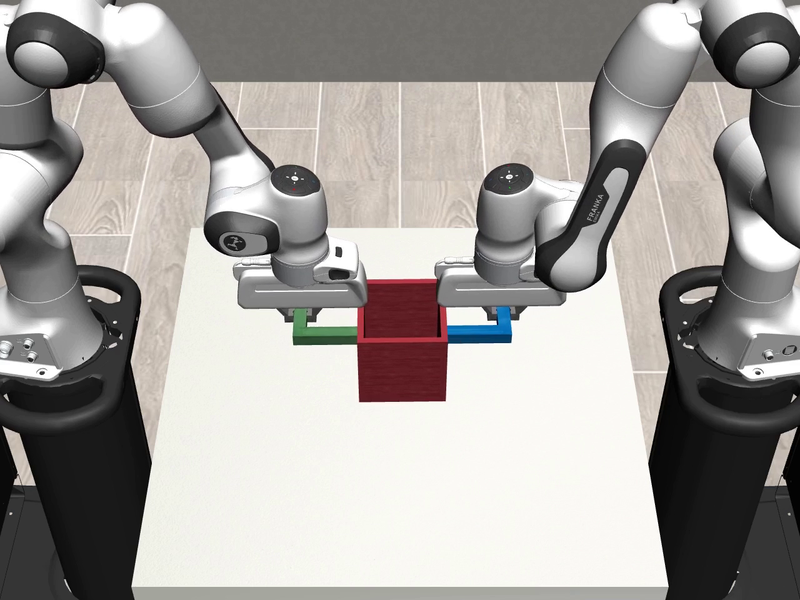}};
    \node[anchor=north west] (ur5e) at ([yshift=-2pt]libero.south west)
      {\includegraphics[trim=120bp 0bp 0bp 0bp,clip,
        width=\setupTileWidth,height=\setupTileHeight]
        {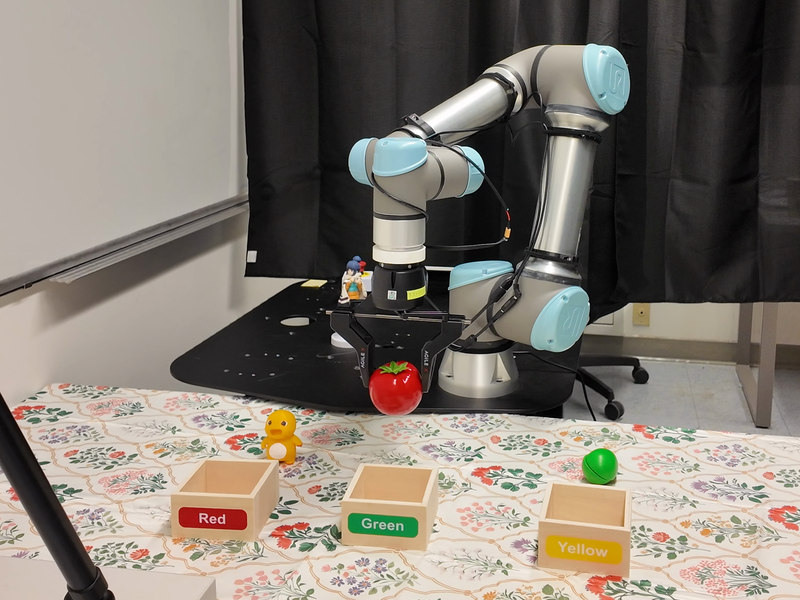}};
    \node[anchor=north west] (wrist) at ([xshift=2pt]ur5e.north east)
      {\includegraphics[width=\setupTileWidth,height=\setupTileHeight]
        {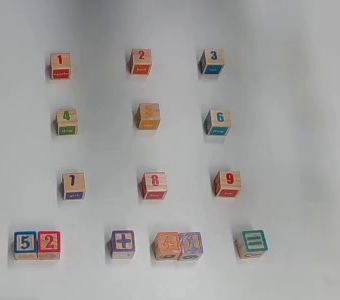}};
    \foreach \panel/\panelletter in {libero/a,robosuite/b,ur5e/c,wrist/d}{
      \node[anchor=north west,fill=white,inner sep=1pt,font=\scriptsize\bfseries]
        at ([xshift=2pt,yshift=-2pt]\panel.north west) {(\panelletter)};
    }
  \end{tikzpicture}\par
  \setlength{\abovecaptionskip}{3pt}
  \caption{\textbf{Experimental setups.}
    \textit{(a)} Representative LIBERO-Pro Long scene.
    \textit{(b)} Representative Robosuite scene.
    \textit{(c)} real front view. \textit{(d)} real top-down view.}
  \label{fig:experimental_setup}
  \vspace{-8pt}
\end{wrapfigure}

To test whether one learning framework can acquire reusable hierarchical task
knowledge across diverse tasks and robot configurations, we train separate
Cerebellum and Brain libraries from scratch on LIBERO-90~\citep{NEURIPS2023_8c3c6668}
and seven Robosuite tasks~\citep{zhu2020robosuite}, using the same two-stage PVG
procedure. We evaluate GPT-5.6 Sol~\citep{openai2026gpt56sol} and Claude Opus
5~\citep{anthropic2026opus5}, learning a separate hierarchy per model and task
set and assigning the same model to all three PVG roles. LIBERO-90 uses a
Franka Panda; Robosuite spans single- and two-arm environments.
In both experiments, the Proposer acquires RGB-D observations on demand
by calling the shared code-based policy interface to request configured camera views,
including top-down views. Task definitions,
robot configurations, and observation settings are detailed in
Appendices~\ref{appendix:libero_protocol} and~\ref{appendix:robosuite_protocol}.\looseness=-1

\subsection{Training Setup}
\begin{figure}[!t]
  \centering
  \setlength{\parskip}{0pt}
  \setlength{\abovecaptionskip}{3pt}
  \includegraphics[width=\linewidth]{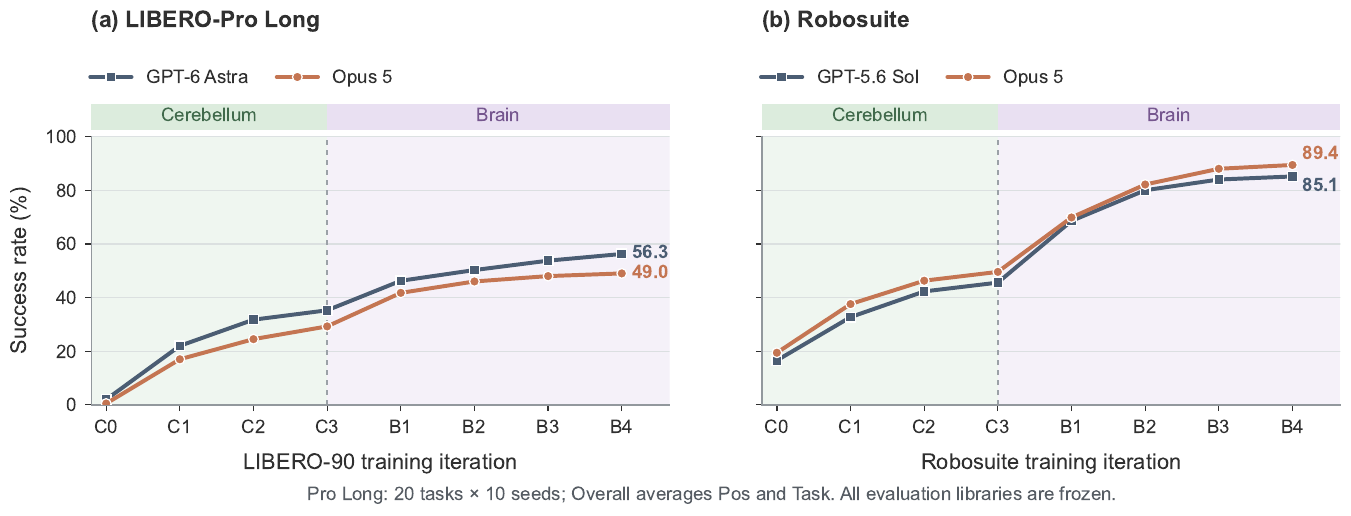}
  \vspace{-0.6cm}
  \caption{\textbf{Skill2Real provides consistent hierarchical gains across
  foundation VLMs.} With GPT-6 Astra, GPT-5.6 Sol, and Opus 5, success rises
  through Cerebellum training and continues through Brain training in both
  settings: \textit{(a)} transfer of LIBERO-90 checkpoints to unseen
  LIBERO-Pro Long tasks; and \textit{(b)} independently learned Robosuite
  libraries. This shared trend demonstrates that the framework's learning
  gains are orthogonal to the underlying VLM.}
  \label{fig:transfer_learning_curve}
\end{figure}

Each benchmark uses three Cerebellum iterations (C1--C3), then four Brain
iterations (B1--B4) with C3 frozen; C0 is evaluation before training. Across
both benchmarks, every iteration covers every task with five seeded training
rollouts per task. This yields 450 episodes per iteration and 3,150 across
C1--B4 on LIBERO-90, and 35 per iteration and 245 total across the seven
Robosuite tasks. Governor-based runs evaluate each candidate over 25 validation
rollouts using seeds disjoint from training. Sol, Opus 5, and the role ablations
share the same per-task training-rollout budget; no-Governor omits admission validation.
Figure~\ref{fig:transfer_learning_curve} reports frozen LIBERO-90 checkpoints
on LIBERO-Pro Long and independently trained Robosuite checkpoints. Additional
evaluations and ablations on the source benchmarks appear in
Appendix~\ref{appendix:experiment_details}.\looseness=-1

\Needspace{4\baselineskip}
\subsection{Real-World Setup}
\label{subsection:real_setup}

All real-world tasks use a single UR5e arm with a Pika gripper and scene
and wrist RGB-D cameras (Appendix~\ref{appendix:real_protocol}).
A robot-specific backend implements the shared
API. The real-world benchmark comprises pick-and-place, attribute-based sorting,
equation assembly, and drawer manipulation
(R1--R4; Appendix~\ref{appendix:real_world_tasks}). Deployment uses no
real-world task demonstrations, task-policy fine-tuning, or online
skill-memory updates.\looseness=-1

\subsection{Evaluation Setup}
\label{subsection:evaluation_setup}

Evaluation isolates the contribution of hierarchical knowledge by comparing
the base Proposer, Cerebellum-only skills, and the full hierarchy, with an
additional Brain-only control on the real robot. On Pro Long, Astra and Opus 5
evaluate each LIBERO-90 checkpoint across 20 tasks and 10 seeds; the Astra
series and training-role ablations use Sol-trained libraries. Real-world and
cross-Proposer comparisons use the Sol-trained C3/B3 checkpoints
(Appendix~\ref{appendix:shared_protocol}). Success denotes complete task
execution, and aggregate scores average per-task success rates. Real-world
results use 20 trials per method for each task (R1--R4); the fixed C3/B3 Pro
Long comparison uses 20 trials per task under each position and instruction
perturbation setting (Appendix~\ref{appendix:long_protocol}).\looseness=-1

\section{Results}
\label{section:experiments}

The evaluation addresses three questions: (1) What do local skill acquisition
in the Cerebellum and task-level composition in the Brain each contribute to
performance? (2) How broadly does the learned executable knowledge transfer
through the shared code-based policy interface across unseen tasks, foundation
VLM Proposers, and real deployment? (3) How much do the Verifier's privileged
diagnosis and the Governor's evidence-based admission contribute to the learned
skill library? We first analyze LIBERO-90-to-Pro Long checkpoint transfer and
independent Robosuite learning for (1), then use real-world and fixed-library
cross-Proposer evaluation for (2), and finally ablate the Verifier and Governor
for (3).
\looseness=-1

\begin{figure}[!t]
  \centering
  \setlength{\parskip}{0pt}
  \setlength{\abovecaptionskip}{3pt}
  \includegraphics[width=\linewidth]{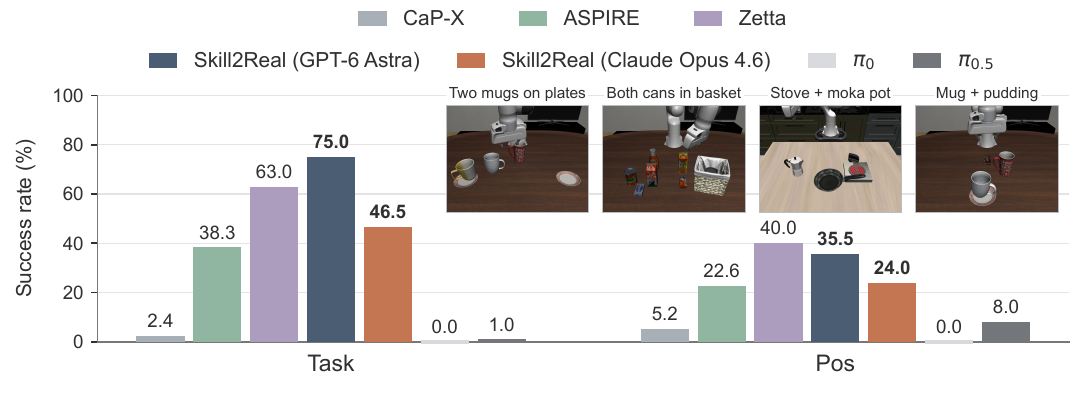}
  \vspace{-0.9cm}
  \caption{\textbf{LIBERO-Pro Long comparison.} Skill2Real with Astra
  achieves the highest Overall and instruction-perturbation (Task) success;
  Zetta leads under position perturbations (Pos). Overall averages Task and
  Pos. Skill2Real evaluates a fixed Sol-trained C3/B3 hierarchy with Astra or
  Opus 4.6. Comparisons include CaP-Agent0~\citep{fu2026capx}, ASPIRE
  ($N{=}90$), $\pi_0$, and $\pi_{0.5}$~\citep{lu2026aspire}, and
  Zetta~\citep{ding2026zetta}.}
  \label{fig:libero_pro_long_comparison}
\end{figure}

\Needspace{4\baselineskip}
\subsection{Skill Transfer Across Training Checkpoints}
\label{subsection:checkpoint_transfer}
\label{subsection:robosuite_learning}

Figure~\ref{fig:transfer_learning_curve}(a) tests whether continued skill
learning on LIBERO-90 improves transfer to unseen long-horizon tasks.
After each GPT-5.6 Sol training iteration on LIBERO-90, we freeze the resulting skill library
and evaluate it on LIBERO-Pro Long with GPT-6 Astra as the Proposer.
Opus 5 is evaluated under the same 20-task, 10-seed target protocol.
Panel (a) therefore traces transfer from successive LIBERO-90 checkpoints to
Pro Long.\looseness=-1

\textbf{LIBERO-90 to LIBERO-Pro Long.} Both stages improve transfer across
Proposers. Cerebellum learning raises Overall success from 2.0\% to 35.3\%
for Astra and from 0.5\% to 29.3\% for Opus 5. With the C3 Cerebellum fixed,
Brain learning further raises success to 56.3\% and 49.0\%, adding 21.0 and
19.8 percentage points. These gains show that the Brain can reuse local skills
learned on LIBERO-90 to construct stronger long-horizon programs for unseen
tasks. At B4, instruction/position success is 76.0\%/36.5\% for Astra and
66.5\%/31.5\% for Opus, indicating that object placement remains more
challenging for both. The source-suite LIBERO-90 curves and
six-family analysis are retained in Appendix~\ref{appendix:libero_protocol};
the checkpoint transfer protocol is in Appendix~\ref{appendix:long_protocol}.\looseness=-1

\textbf{Robosuite.} The same two-stage benefit appears when each model learns
its own hierarchy from scratch across single-arm and bimanual Robosuite
tasks~\citep{zhu2020robosuite} (Figure~\ref{fig:transfer_learning_curve}(b)).
Cerebellum learning raises seven-task mean success from 16.6\% to 45.6\% for
Sol and from 19.4\% to 49.6\% for Opus. Brain learning then composes the fixed
C3 skills to reach 85.1\% and 89.4\%, adding 39.6 and 39.9 percentage points
and improving all seven tasks.\looseness=-1

Across both benchmarks and all three Proposers, the two stages play
complementary roles: the Cerebellum acquires reusable manipulation skills,
and the Brain composes those fixed skills into more capable task programs.
Concrete Cerebellum and Brain memory examples are shown in
Appendix~\ref{appendix:pvg_skill_examples}.

\Needspace{13\baselineskip}
\subsection{Zero-Shot Real-World Transfer}
\label{subsection:real_world_transfer}

\begin{wraptable}[11]{r}{0.48\textwidth}
  \vspace{-\intextsep}
  \vspace{-\abovecaptionskip}
  \caption{\textbf{Real-world task completion} (\%; 20 trials per
  method/task). R1: pick-and-place; R2: sorting; R3: equation assembly;
  R4: drawer manipulation. Mean is the unweighted average across R1--R4.}
  \label{tab:real_results}
  \centering\small
  \setlength{\tabcolsep}{2.5pt}
  \begin{tabular}{@{}lrrrrr@{}}
    \toprule
    Method & R1 & R2 & R3 & R4 & Mean \\
    \midrule
    CaP-Agent0 (CaP-X) & 15 & 20 & 10 & 0 & 11.25 \\
    Astra, no learned skills & 40 & 35 & 35 & 0 & 27.50 \\
    Astra, Brain only & 50 & 60 & 40 & 0 & 37.50 \\
    Astra, Cerebellum only & 80 & 80 & 65 & 0 & 56.25 \\
    \textbf{Astra, full Skill2Real} & \textbf{95} & \textbf{85} & \textbf{85} & \textbf{50} & \textbf{78.75} \\
    \bottomrule
  \end{tabular}
\end{wraptable}

With the Astra Proposer and API held fixed, the full hierarchy raises mean
real-world completion from 27.50\% without learned skills to 78.75\%
(Table~\ref{tab:real_results}), outperforming both the Brain-only (37.50\%)
and Cerebellum-only (56.25\%) variants. Adding the Brain to the Cerebellum
improves R1--R3 by 15, 5, and 20 percentage points and is the only
configuration to complete drawer manipulation (50\%). The hierarchy therefore
contributes beyond reliable local execution: task-level composition turns
reusable operations into stronger multi-step behavior.\looseness=-1

Figures~\ref{fig:sim_real_skills} and~\ref{fig:real_execution} show the learned
local skills and their composition in recorded real executions.
Appendix~\ref{appendix:real_world_tasks} provides additional examples under
varied appearance and lighting; Appendix~\ref{appendix:real_protocol} gives
the evaluation protocol.\looseness=-1

\subsection{Zero-Shot Long-Horizon Generalization}
\label{subsection:libero_pro_long}

Figure~\ref{fig:libero_pro_long_comparison} compares Skill2Real with
CaP-Agent0~\citep{fu2026capx}, ASPIRE~\citep{lu2026aspire}, and
Zetta~\citep{ding2026zetta} on LIBERO-Pro Long. Using Astra with the fixed
Sol-trained C3/B3 hierarchy, Skill2Real attains the best Overall (55.3\%) and
Task (75.0\%) success, exceeding Zetta's 51.5\% and 63.0\%. The larger Task
margin reflects the value of hierarchical executable knowledge: local
manipulation skills remain reusable while Brain programs rebind and compose
them for redirected instructions. The same hierarchy reaches 35.3\% Overall
with Opus 4.6, above ASPIRE's 30.5\%, showing that the learned knowledge also
transfers across test-time Proposers.\looseness=-1

Zetta leads under position perturbations (40.0\% versus 35.5\%). This gap is
driven by cases such as bowl-in-drawer and book-in-caddy, where an end-to-end
VLA policy provides tighter reactive control for narrow-space pick-and-place
than the current code-based skill interface. Appendix~\ref{appendix:long_protocol}
provides the per-task breakdown.\looseness=-1

\Needspace{22\baselineskip}
\subsection{Verifier and Governor Ablations}
\label{subsection:ablations}

\begin{wrapfigure}[18]{r}{0.48\textwidth}
  \vspace{-\intextsep}
  \centering
  \setlength{\abovecaptionskip}{3pt}
  \includegraphics[width=\linewidth]{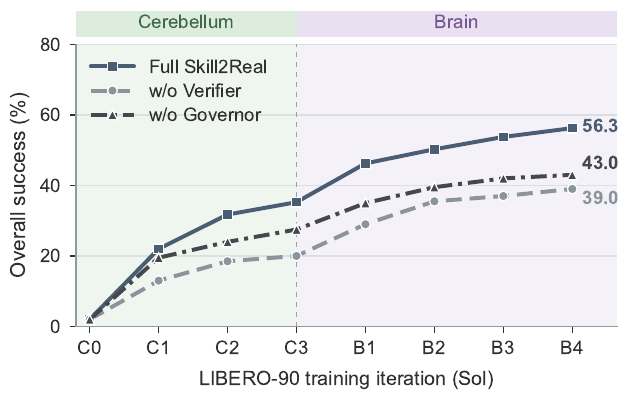}
  \caption{\textbf{Both PVG roles improve transfer.} Removing either role
  lowers Pro Long success throughout hierarchical learning; labels give B4
  Overall success (\%).}
  \label{fig:pvg_pro_long_curve}
\end{wrapfigure}

PVG separates two complementary forms of learning-time supervision: the
Verifier diagnoses execution from privileged simulation evidence, and the
Governor admits validated updates. We ablate each role during LIBERO-90
training to test how these functions shape transfer to Pro Long
(Figure~\ref{fig:pvg_pro_long_curve}).

Full Skill2Real leads at every iteration and reaches 56.3\% Overall, versus
39.0\% without the Verifier and 43.0\% without the Governor---losses of 17.3
and 13.3 percentage points. The gap appears during Cerebellum learning and
persists as the Brain composes the fixed C3 library, showing that grounded
feedback and selective memory admission jointly improve transferable skill
learning at both hierarchy levels. Appendix~\ref{appendix:pvg_protocol} gives
the ablation protocol and source-suite LIBERO-90 results.\looseness=-1

\section{Conclusion and Discussion}
\label{section:conclusion_and_discussion}

Skill2Real defines sim-to-real transfer at the level of executable task
knowledge.  Its multi-agent learning system uses privileged simulation
evidence to build a two-level hierarchy: the Cerebellum captures reusable
local manipulation skills, and the Brain composes them into long-horizon task
programs.  Expressing both levels through a shared code-based policy interface
allows the frozen hierarchy to be grounded in real observations by a
foundation VLM.\looseness=-1

Across LIBERO-90, Robosuite, LIBERO-Pro Long, and UR5e experiments, the two
levels provide complementary capabilities: local skills improve reliable
interaction, while task-level composition supports reuse on unseen and
long-horizon tasks.  In particular, successive frozen LIBERO-90 checkpoints
raise Astra's success on unseen LIBERO-Pro Long from 2.0\% to 56.3\%, while
the full hierarchy reaches 78.75\% mean completion on UR5e versus 56.25\% for
the Cerebellum alone.  The Proposer--Verifier--Governor ablations further show
that privileged diagnosis and evidence-based memory admission materially
improve skill acquisition.  Together, these results demonstrate that
simulation-learned programs can generalize across tasks, scenes, and the real
robot while their learned hierarchy remains fixed.\looseness=-1

Skill2Real currently operates through a shared policy interface whose robot
backend provides perception and low-level control.  Future work will extend
the current pure code-as-policy interface by exposing end-to-end policies as
callable tools, and reduce VLM inference latency for faster closed-loop
execution.\looseness=-1

\FloatBarrier

\clearpage
\section*{AI Disclosure}

We used generative AI tools to assist with manuscript drafting and language
polishing. The LLMs used within our proposed method and their roles are described
in the main text. The authors reviewed and revised all AI-assisted text to ensure
accuracy, clarity, and consistency with the research conducted. We take
responsibility for the final content of this work, including text, claims or
artifacts produced with the aid of generative AI.

\clearpage
\appendix
\section{Real-World Tasks and Visual Generalization}
\label{appendix:real_world_tasks}
\setcounter{table}{0}
\setcounter{figure}{0}
\renewcommand{\thetable}{\thesection.\arabic{table}}
\renewcommand{\theHtable}{appendix.\thesection.\arabic{table}}
\renewcommand{\thefigure}{\thesection.\arabic{figure}}
\renewcommand{\theHfigure}{appendix.\thesection.\arabic{figure}}

\begin{table}[H]
  \caption{Real-world task suite. R1--R4 are the task identifiers used in
  Table~\ref{tab:real_results}; object instances are specified per trial.}
  \label{tab:real_task_suite}
  \centering
  \small
  \setlength{\tabcolsep}{3pt}
  \begin{tabular}{@{}l>{\raggedright\arraybackslash}p{0.21\linewidth}
      >{\raggedright\arraybackslash}p{0.31\linewidth}
      >{\raggedright\arraybackslash}p{0.33\linewidth}@{}}
    \toprule
    ID & Task & Instruction template & Completion condition \\
    \midrule
    R1 & Pick-and-place & Place the specified object in the specified container.
       & The target object is inside the target container. \\
    R2 & Attribute-based sorting & Sort objects by color or semantic category.
       & Each object is placed at the destination assigned to its category. \\
    R3 & Equation assembly & Move cubes to form the requested equation.
       & The cubes form the requested equation in the correct order. \\
    R4 & Drawer manipulation & Open the target drawer, insert the specified object, then close the drawer.
       & The target object is inside the correct drawer, which is closed. \\
    \bottomrule
  \end{tabular}
\end{table}

\begin{figure}[H]
  \centering
  \newlength{\realtasktilewidth}
  \setlength{\realtasktilewidth}{\dimexpr(\linewidth-8pt)/2\relax}
  \begin{tabular}{@{}c@{\hspace{8pt}}c@{}}
    \includegraphics[width=\realtasktilewidth]{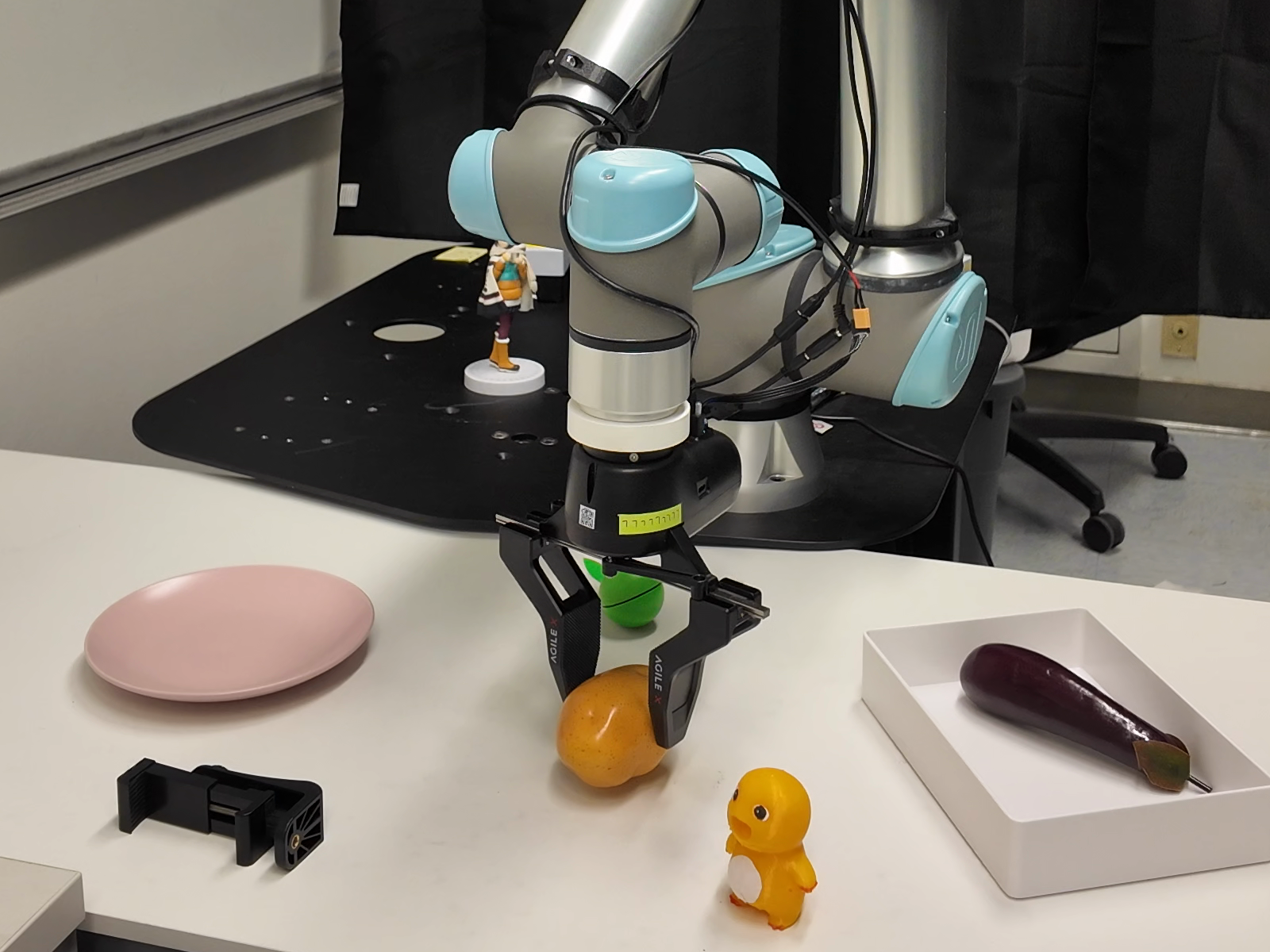}
    & \includegraphics[width=\realtasktilewidth]{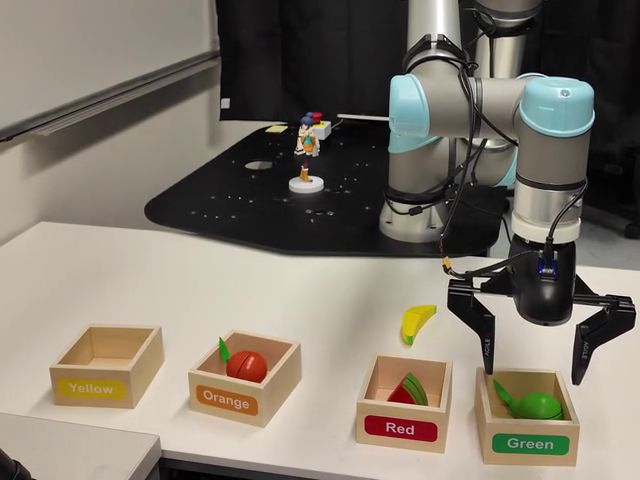} \\[2pt]
    {\small\textbf{(a) R1: Pick-and-place}}
    & {\small\textbf{(b) R2: Attribute-based sorting}} \\[7pt]
    \includegraphics[width=\realtasktilewidth]{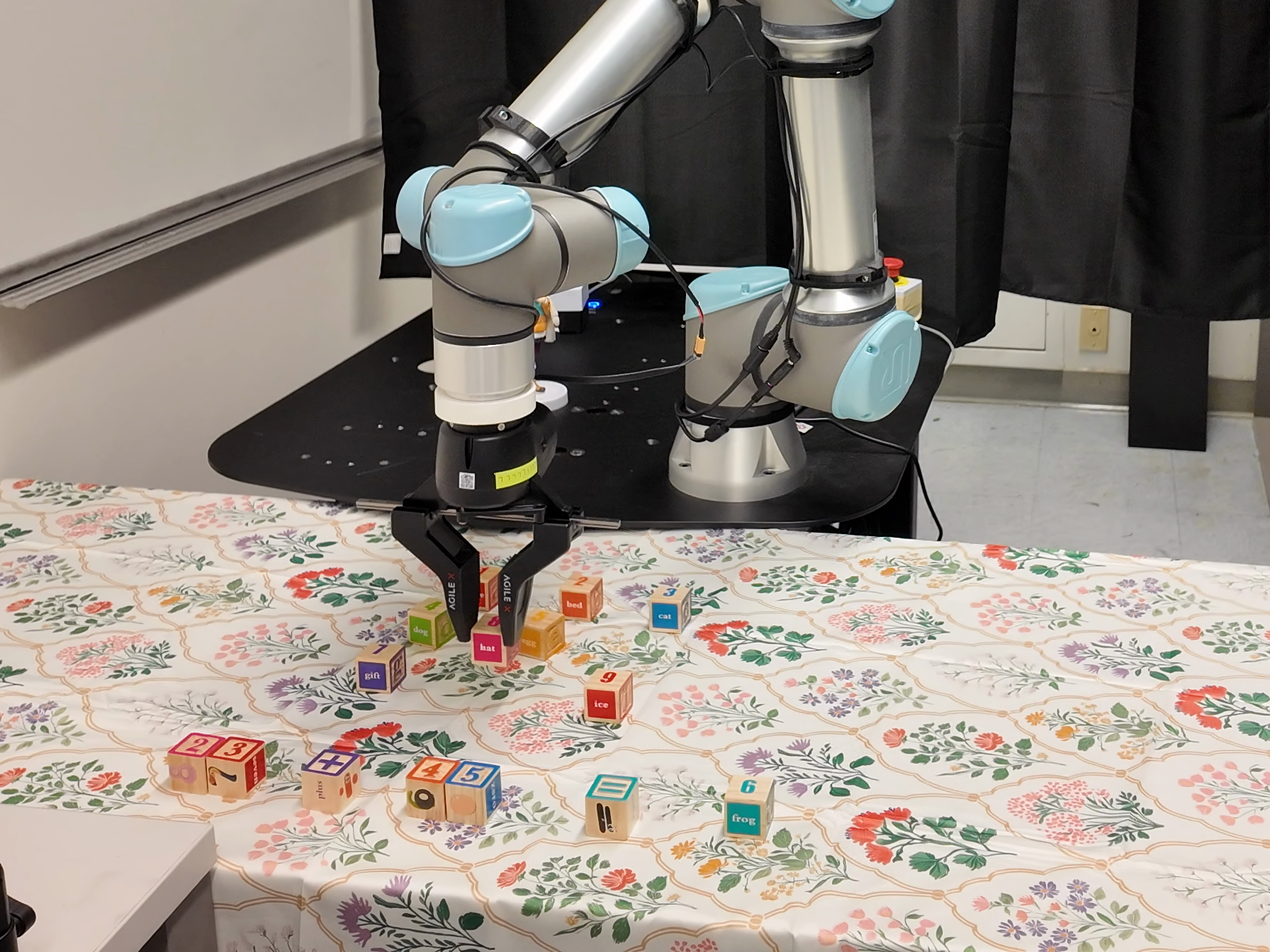}
    & \includegraphics[width=\realtasktilewidth]{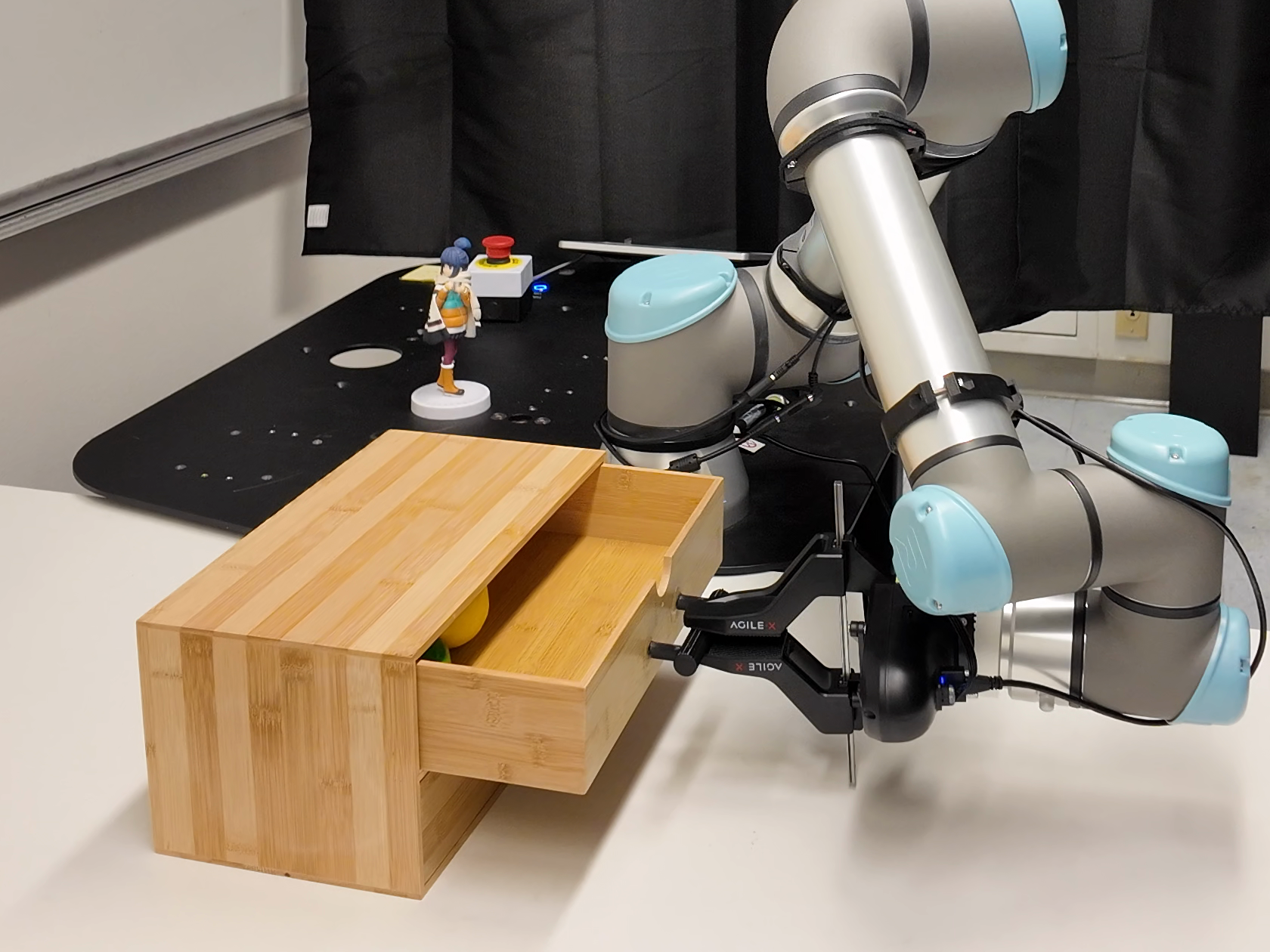} \\[2pt]
    {\small\textbf{(c) R3: Equation assembly}}
    & {\small\textbf{(d) R4: Drawer manipulation}}
  \end{tabular}
  \caption{\textbf{Representative real-world task scenes.}
  One recorded image illustrates each task in Table~\ref{tab:real_task_suite}:
  (a) picking an object for placement on a plate;
  (b) sorting objects into color-labeled boxes;
  (c) arranging answer cubes for $23+45=68$; and
  (d) drawer manipulation.}
  \label{fig:real_task_examples}
\end{figure}

\Needspace{8\baselineskip}
\subsection{Tabletop Appearance and Lighting Variation}
\label{appendix:visual_generalization}

\begin{figure}[H]
  \centering
  \newlength{\realvariationtilewidth}
  \setlength{\realvariationtilewidth}{\dimexpr(\linewidth-8pt)/2\relax}
  \begin{tabular}{@{}c@{\hspace{8pt}}c@{}}
    \includegraphics[width=\realvariationtilewidth]{figures/real_appendix/task_a_pick_place.png}
    & \includegraphics[width=\realvariationtilewidth]{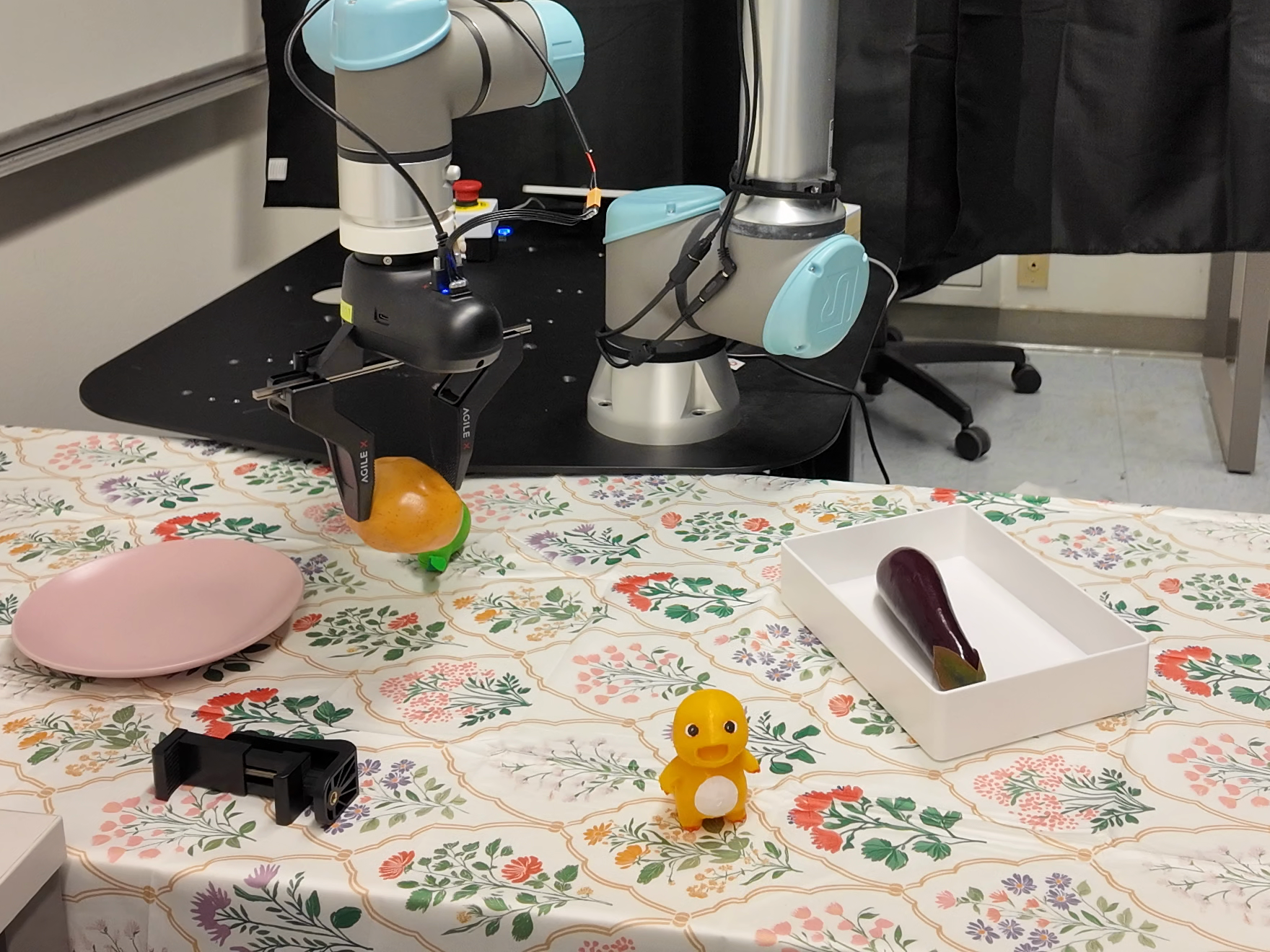} \\[2pt]
    {\small\textbf{(a) White tabletop}}
    & {\small\textbf{(b) Floral cloth}} \\[7pt]
    \includegraphics[width=\realvariationtilewidth]{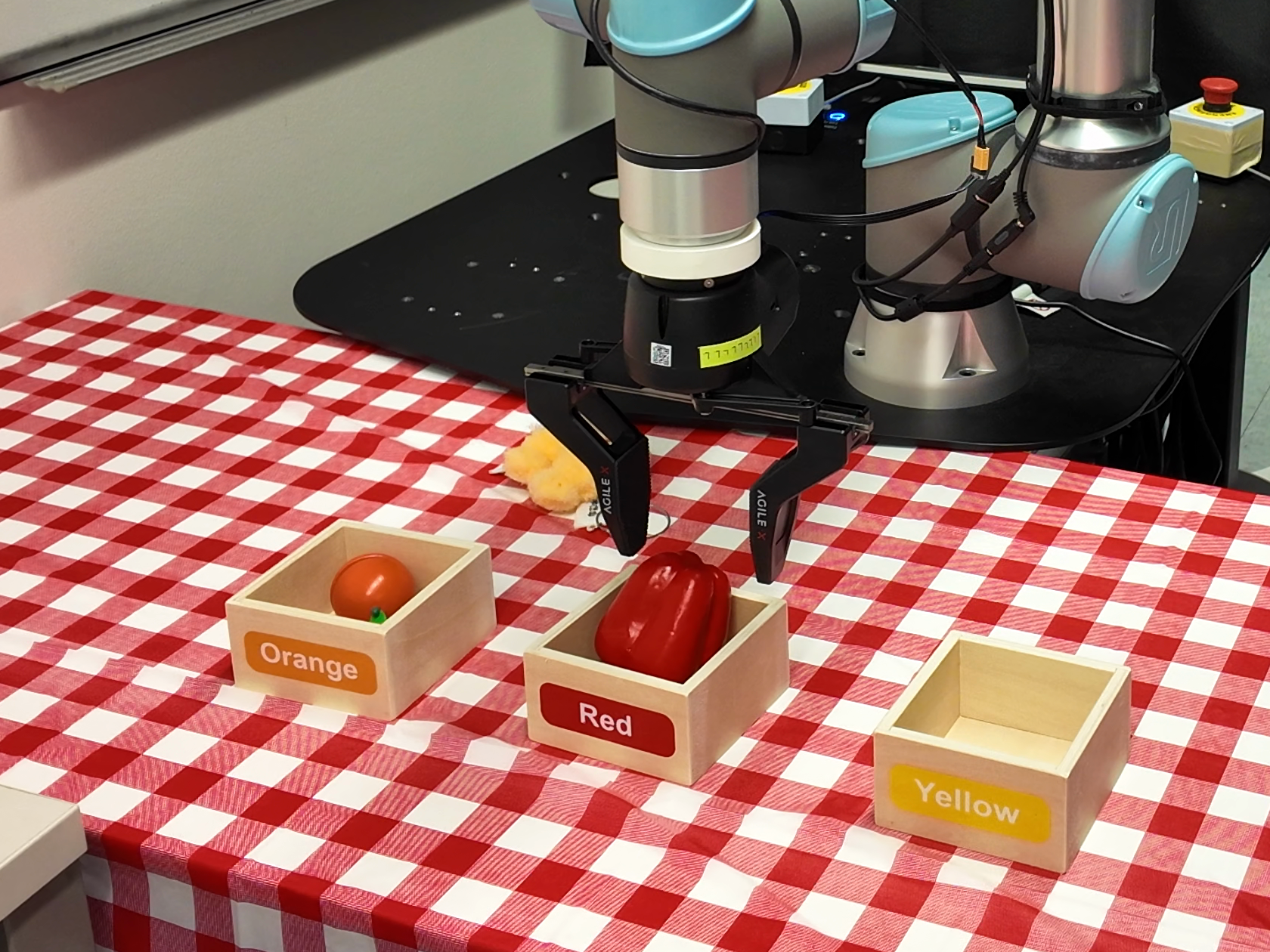}
    & \includegraphics[width=\realvariationtilewidth]{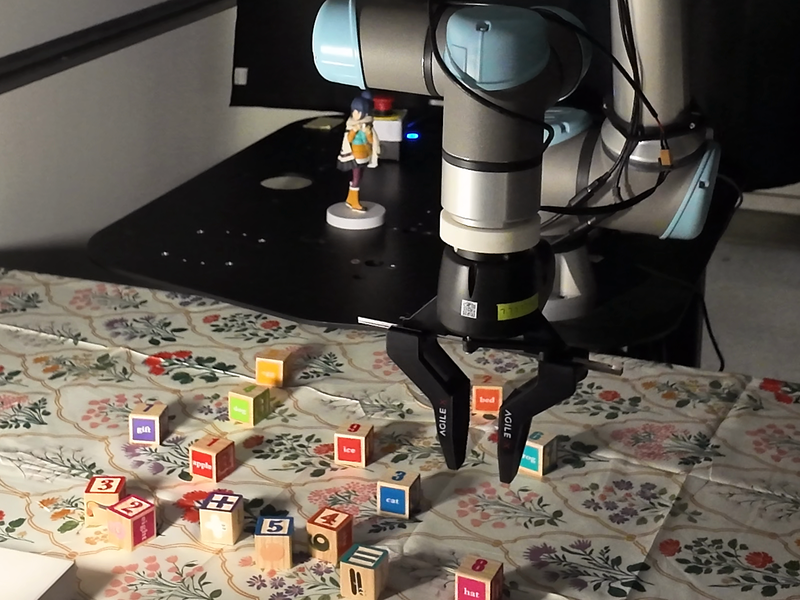} \\[2pt]
    {\small\textbf{(c) Red gingham}}
    & {\small\textbf{(d) Dim side lighting}}
  \end{tabular}
  \caption{\textbf{Tabletop and lighting variation in real deployment.}
  Recorded scenes span white, floral, and red-gingham surfaces and
  dim side lighting.
  All images retain their recorded exposure; the examples are qualitative.}
  \label{fig:real_visual_generalization}
\end{figure}

\paragraph{Generalization across tabletop appearance.}
The recorded deployments provide qualitative evidence that the learned
manipulation skills can be reused across changes in tabletop color and
texture (Figure~\ref{fig:real_visual_generalization}).
The white-table and floral-cloth examples show the same potato-to-plate
operation, with task completion confirmed in the archived reviews for
both backgrounds. On the red gingham cloth, the robot executes three
object transfers into color-labeled boxes; an external-video review
confirms all three placements. Together, these examples support reuse
across visually distinct real scenes. They complement the aggregate
task-completion results in Table~\ref{tab:real_results}.

\paragraph{Execution under changed lighting.}
Under dim, uneven side lighting, the robot executes two answer-cube
pick-and-place sequences, including both releases and the return motion
(panel (d)); the record ends before final equation acceptance.
This qualitative example does not supply a separate condition-wise success
rate or add trials to Table~\ref{tab:real_results}.

\FloatBarrier
\section{Additional Experimental Details}
\label{appendix:experiment_details}
\setcounter{table}{0}
\setcounter{figure}{0}
\renewcommand{\thefigure}{\thesection.\arabic{figure}}
\renewcommand{\theHfigure}{appendix.\thesection.\arabic{figure}}

\subsection{Shared Configuration and Evaluation Protocol}
\label{appendix:shared_protocol}

\paragraph{Models and training iterations.}
We train separate LIBERO-90 libraries with GPT-5.6 Sol~\citep{openai2026gpt56sol}
and Claude Opus 5~\citep{anthropic2026opus5}. Within
each training run, the Proposer, Verifier, and Governor use the same model,
with reasoning effort set to \texttt{medium}. C0 denotes the no-skill
baseline. C1--C3 denote the first through third cerebellum training
iterations; B1--B4 denote the first through fourth brain training iterations with the C3
cerebellum library held fixed. For the Astra Pro Long checkpoint series and
role ablations, GPT-5.6 Sol trains the skills on LIBERO-90 and GPT-6 Astra
evaluates each frozen checkpoint on LIBERO-Pro Long. The checkpoint curves
also include Opus 5 under the same 20-task, 10-seed evaluation protocol.
No Pro Long task is
used for training or skill updates. The separate real-world and
cross-Proposer comparisons in Table~\ref{tab:real_results}
and Figure~\ref{fig:libero_pro_long_comparison} retain Sol's LIBERO-90 cerebellum library at
C3 and brain library at B3. These fixed libraries are shared across the
test-time Proposers in those comparisons. Robosuite libraries
are trained independently and are excluded from these transfer evaluations.

\paragraph{Controlled evaluation.}
Both memories are frozen before each held-out evaluation; the Proposer
receives public visual observations and API returns, without privileged
simulator state or feedback from the Verifier and Governor. Within our
memory-stage and role comparisons for a fixed Proposer, the API
documentation, task definitions, and evaluation instances are held fixed.
Comparisons across Proposers and with published methods use the models
and protocols specified in the corresponding subsections. Success refers
to complete tasks. Aggregate scores average per-task success rates, and
missing results are marked with dashes rather than treated as zero success.

\paragraph{Reporting scope.}
Success rates are point estimates for the recorded evaluations of fixed
libraries. Repeated episodes assess a fixed library's performance; they
do not measure variability across independent training runs. Learning
curves are indexed by training iteration, not cumulative rollout cost.
Stage comparisons track the combined effect of additional training and
the second memory level, rather than isolating hierarchy at a matched
training budget.

\subsection{LIBERO-90 Families and Stage Comparisons}
\label{appendix:libero_protocol}
\label{subsection:libero90_learning}

\paragraph{Robot and observations.}
LIBERO uses a seven-degree-of-freedom Franka Panda arm with a gripper and
joint-position control at 20\,Hz, with a maximum episode horizon of 4,000
environment steps. The Proposer obtains visual observations on demand
by calling the shared code-based policy interface. The configured views include a fixed
scene camera and a wrist-mounted camera, each providing RGB images at
\mbox{$800\times512$} resolution; top-down views can also be requested
through the same API. Metric depth is reconstructed from calibrated
top-down stereo RGB images through the documented depth operation. An
image-path return alone does not provide depth or calibration; the wrist
operation includes these only where documented and available. End-effector
pose is available through its public query. Joint-position control does not
imply a public joint-state query. Privileged simulator evidence is restricted
to execution evaluation and the learning-time PVG loop.

\paragraph{Evaluation instances and aggregation.}
The LIBERO-90 analysis asks which types of manipulation benefit from local
skills and which improve further through task-level composition.  Cerebellum
and brain training have been completed with Sol and Opus as the Proposer.
We organize evaluation by six task families and compare three
memory states: no learned skills, the frozen Stage-I cerebellum, and the full
hierarchy after Stage II. Evaluation at C0 and after each training iteration
uses the same five held-out initial states per task, giving 450 trials per
evaluation. Overall success
is the number of successful trials divided by 450. Family success is the
unweighted mean of its task success rates; the suite-level score averages
over all 90 tasks, rather than giving differently sized families equal weight.

\begin{figure}[!htbp]
  \centering
  \includegraphics[width=\linewidth]{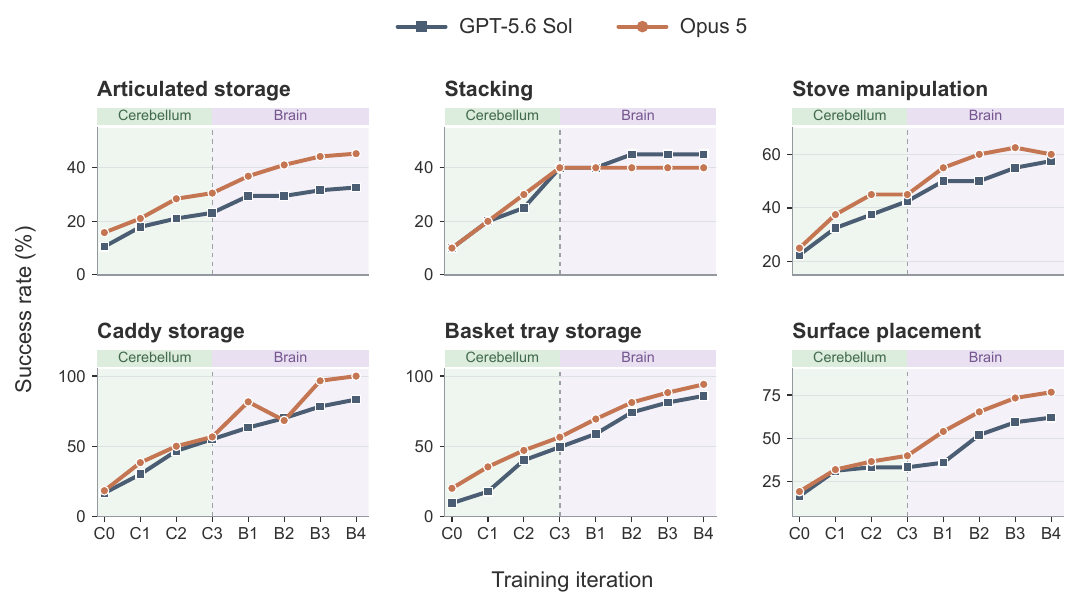}
  \caption{\textbf{Learning across six LIBERO-90 families.}
  Each family is normalized by its full mark and uses its own vertical scale.
  C0 is the no-skill baseline; C1--C3 track cerebellum learning, and B1--B4
  track subsequent brain learning. The background regions meet at C3.}
  \label{fig:libero90_family_learning}
\end{figure}

Table~\ref{tab:libero90_families} describes the manipulation scope of each
family. These are analysis groups within LIBERO-90~\citep{NEURIPS2023_8c3c6668},
rather than separate benchmark suites. The row order matches the panels in
Figure~\ref{fig:libero90_family_learning}.

\begin{table}[!htbp]
  \caption{\textbf{Six task families in the LIBERO-90 analysis.}
  Full mark is the score denominator used to normalize each family's
  learning curve, not its task count. The six denominators sum to 450.}
  \label{tab:libero90_families}
  \centering\small
  \setlength{\tabcolsep}{5pt}
  \renewcommand{\arraystretch}{1.15}
  \begin{tabular}{@{}>{\raggedright\arraybackslash}p{0.23\linewidth}
      >{\raggedright\arraybackslash}p{0.60\linewidth}r@{}}
    \toprule
    Family & Manipulation scope and representative operations & Full mark \\
    \midrule
    Articulated storage & Opening or closing cabinet drawers and microwave
      doors; placing objects inside drawers. & 95 \\
    Stacking & Placing one bowl on another, requiring alignment and stable
      support after release. & 20 \\
    Stove manipulation & Switching a stove on or off; placing frying pans
      or moka pots on the stove. & 40 \\
    Caddy storage & Placing objects into specified caddy compartments,
      requiring compartment selection and constrained placement. & 60 \\
    Basket / tray storage & Transferring household objects into open baskets
      or trays. & 85 \\
    Surface placement & Placing objects on plates, shelves, or cabinet tops,
      or at specified positions relative to other objects. & 150 \\
    \bottomrule
  \end{tabular}
\end{table}

\begin{figure}[!htbp]
  \centering
  \includegraphics[width=0.9\linewidth]{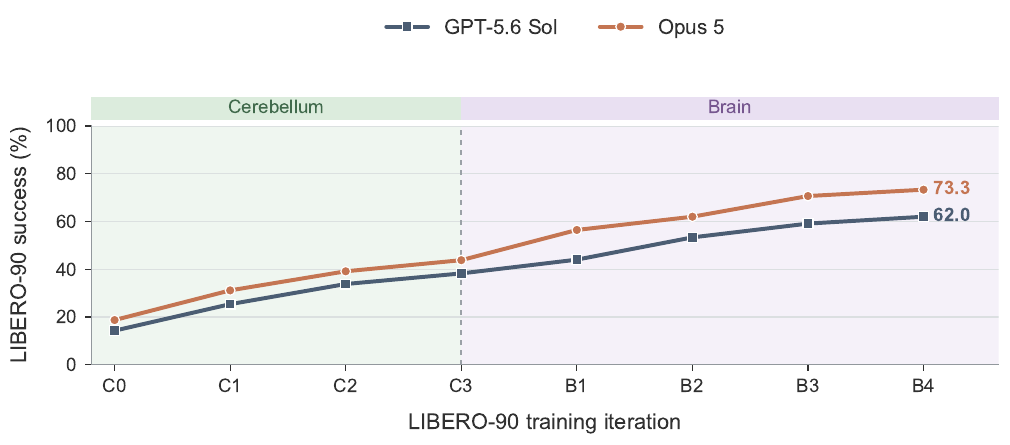}
  \caption{\textbf{Source-suite skill learning on LIBERO-90.}
  Sol and Opus train separate libraries and are evaluated on the same
  450 held-out trials after each iteration. C0 is the no-skill baseline;
  C1--C3 and B1--B4 denote Cerebellum and Brain training. The Cerebellum
  remains fixed after C3, and both memories are frozen during evaluation.}
  \label{fig:libero90_learning_curve}
\end{figure}

Figure~\ref{fig:libero90_learning_curve} shows overall performance;
Figure~\ref{fig:libero90_family_learning} reports the six families using
independent vertical scales. Sol and Opus are distinguished by color;
the cerebellum and brain background regions meet at C3.
The additional brain stage produces larger gains in caddy storage,
basket/tray storage, and surface placement than in stacking for both
models. Opus reaches 100.0\% in caddy storage at B4, up from 56.7\% at
C3, while its stacking score remains at 40.0\%. Thus, the aggregate gain
does not imply equal improvement across manipulation types. These are
independently trained libraries for each Proposer; reuse of the same
frozen Sol-trained libraries across Proposers is evaluated separately in
Appendix~\ref{appendix:long_protocol}.

\paragraph{Overall source-suite learning.}
Sol improves from 14.2\% at C0 to 38.2\% at C3 and 62.0\% at B4;
Opus improves from 18.7\% to 43.8\% and 73.3\%, respectively.
Brain learning adds 23.8 and 29.6 percentage points with the C3
Cerebellum fixed. These curves evaluate the independently trained models
on LIBERO-90; the main-text curves instead evaluate frozen checkpoints
on Pro Long using Astra and Opus 5.

\subsection{Independent Robosuite Training}
\label{appendix:robosuite_protocol}

\paragraph{Tasks, robots, and observations.}
We report seven Robosuite tasks: cube lifting, stacking, and restacking;
nut assembly; spill wiping; and two-arm lifting and handover. The configured
task set also includes pick-and-place with a can, milk carton, or cereal
box; these three tasks have no reported results and are excluded from the
mean. Single-arm tasks use
one manipulator, while lifting and handover use two arms. Robosuite allows
the robot model to be selected through the environment configuration.
Both configurations use joint-position control through the shared code-based
policy interface. The Proposer acquires RGB-D observations through this interface to
request the configured camera views. Task-facing images are rendered at
\mbox{$512\times512$} resolution, using a robot-view camera for single-arm
tasks and a scene camera for bimanual tasks. Configured top-down views
are accessible through the same observation API.

\paragraph{Independent acquisition and reporting.}
The Robosuite experiment examines whether the same two-stage learning
procedure is effective on another set of manipulation
tasks~\citep{zhu2020robosuite}.  Skills are trained anew on Robosuite, using its
backend for the shared interface, rather than initialized from the LIBERO-90
memories. We report separate Sol and Opus runs at C0 and after each of three
cerebellum iterations and four brain iterations, holding the C3 cerebellum
fixed throughout brain learning. Both memories are frozen during evaluation.
Mean success gives equal weight to the seven reported tasks. Since LIBERO itself uses
Robosuite, this experiment broadens task-suite coverage within the simulation
framework; it does not constitute transfer to a different physics engine.

\begin{figure}[!htbp]
  \centering
  \includegraphics[width=\linewidth]{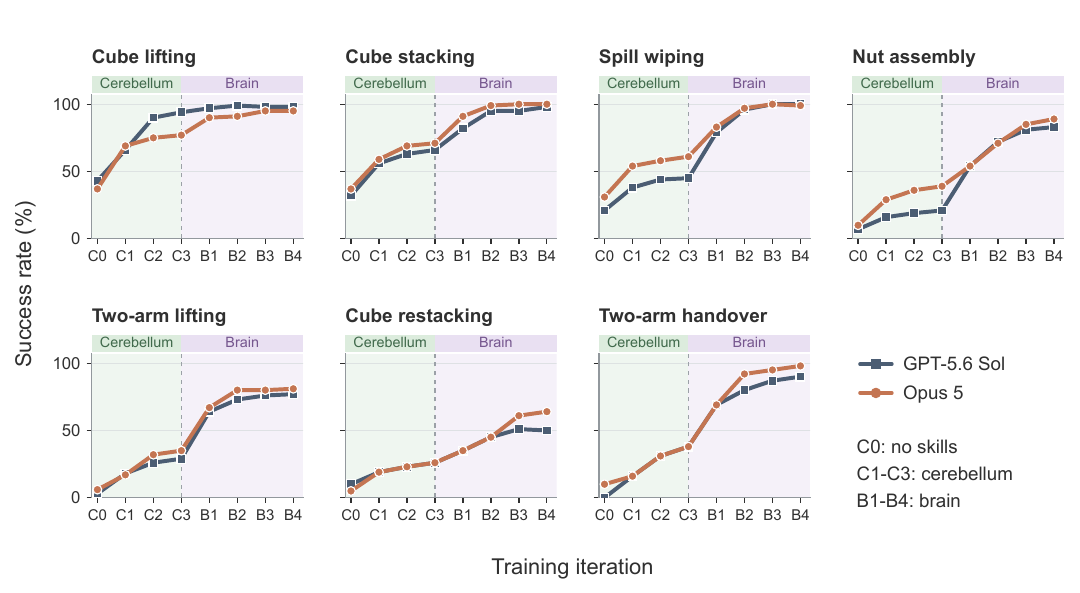}
  \caption{\textbf{Learning across seven Robosuite tasks with Sol and Opus.}
  Task success uses a shared vertical scale. C0 is the no-skill baseline;
  C1--C3 track cerebellum learning, and B1--B4 track brain learning with the
  C3 cerebellum fixed. Background regions meet at C3.}
  \label{fig:robosuite_task_learning}
\end{figure}

\paragraph{Task-level progress.}
Brain-stage gains occur on every reported task
(Figure~\ref{fig:robosuite_task_learning}).
From C3 to B4, nut assembly improves by 62 and 50 percentage points for
Sol and Opus, respectively; two-arm lifting improves by 48 and 46 points,
and handover by 52 and 60 points. Cube lifting improves from 94\% to 98\%
for Sol and from 77\% to 95\% for Opus. Thus, the aggregate progress
includes substantial gains in assembly and bimanual manipulation.

\paragraph{Final checkpoint results.}
At B4, Opus leads on five of seven tasks, while Sol scores higher on cube
lifting (98\% versus 95\%) and spill wiping (100\% versus 99\%).

\subsection{Long-Horizon Transfer Protocol and Per-Task Results}
\label{appendix:long_protocol}

LIBERO-Pro Long tests whether skills learned on shorter source tasks can be
composed on unseen long-horizon tasks.  For this experiment, Skill2Real
memories are learned only on LIBERO-90 and frozen before evaluation; the
independently trained Robosuite memories are not used.  Neither LIBERO-Long
nor LIBERO-Pro Long tasks are used for skill construction, prompt examples,
debugging, bootstrapping, or memory updates. Overall success is the equal
mean of the position-perturbation (Pos) and instruction-perturbation (Task)
success rates.

\paragraph{Evaluation of source-training checkpoints.}
Figure~\ref{fig:transfer_learning_curve}(a) evaluates GPT-6 Astra and
Opus 5 on LIBERO-Pro Long with 20 tasks and 10 seeds per task.
For the Astra series, GPT-5.6 Sol performs skill training on LIBERO-90;
after each iteration, the resulting library is frozen for target evaluation.
C0 has no learned skills; C1--C3 use the corresponding Cerebellum library,
and B1--B4 pair the frozen C3 Cerebellum with each successive Brain library.
The target evaluations do not contribute training rollouts, feedback, or
updates to later checkpoints. Figure~\ref{fig:transfer_learning_curve}(a)
shows Overall success for both evaluation models across this series.

\paragraph{Separate fixed-library cross-Proposer comparison.}
The comparison below retains the previously recorded C3/B3 evaluation,
with the same Sol-trained libraries shared by Astra and Opus 4.6.
It is a separate evaluation record from the C0--B4 checkpoint series;
its aggregate and per-task results are not substituted into that curve.
This earlier record uses ten tasks with 20 trials per task under each
perturbation setting, with success macro-averaged over tasks.

Figure~\ref{fig:libero_pro_long_comparison} compares Skill2Real using
GPT-6 Astra~\citep{openai2026gpt6astra} and
Claude Opus 4.6~\citep{anthropic2026opus46} against ASPIRE~\citep{lu2026aspire}
and CaP-Agent0 from CaP-X~\citep{fu2026capx} using Claude Opus 4.6,
and Zetta~\citep{ding2026zetta}. The ASPIRE and CaP-Agent0 scores
are reported by \citet{lu2026aspire}.
ASPIRE uses its full $N{=}90$ LIBERO-90 skill library and generates one
program per target task, evaluated on seeds 1--50 without additional
debugging, retries, or task-specific library updates. CaP-Agent0 uses
per-seed program generation with test-time reasoning and retries.
For Skill2Real, evaluation excludes target-task skill updates and access
to the privileged Verifier or Governor.

Zetta~\citep{ding2026zetta} evolves a harness around a frozen
$\pi_{0.5}$ action policy using 50 development seeds per target task,
excluding test seeds 1--20. Its final evaluation uses those 20 held-out
seeds. Its LIBERO-10 (Long) position-swap setting S and
instruction-redirection setting T correspond to Pos and Task,
respectively. Zetta's 40.0\% Pos and 63.0\% Task scores give 51.5\% Overall.

Tables~\ref{tab:long_per_task_pos} and~\ref{tab:long_per_task_task}
provide the per-task breakdown for Skill2Real, ASPIRE, and Zetta.
CaP-Agent0 is omitted because only its aggregate scores are available.

Skill2Real's success depends on the perturbation and test-time Proposer.
For example, Astra achieves 90.0\% on both the two-mug and mug-plus-pudding
tasks under instruction perturbations, but 35.0\% and 25.0\% under position
perturbations. Neither Proposer succeeds on the soup-plus-cream-cheese
task under position perturbations. The Pos--Task gap also varies across
tasks: on soup-plus-tomato-sauce, Astra instead scores 80.0\% on Pos and
30.0\% on Task. Position perturbations are therefore harder in aggregate,
with substantial variation across individual tasks.

\begin{table}[!htbp]
  \caption{\textbf{Per-task LIBERO-Pro Long success under position
  perturbations (Pos).} Success rates (\%) for the fixed C3/B3 comparison.
  Skill2Real uses 20 trials
  per task. Baselines: ASPIRE ($N{=}90$)~\citep{lu2026aspire};
  Zetta (S)~\citep{ding2026zetta}.
  $\pi_{0.5}$ is Zetta's base policy.}
  \label{tab:long_per_task_pos}
  \centering\small
  \setlength{\tabcolsep}{4pt}
  \renewcommand{\arraystretch}{1.1}
  \begin{tabular*}{\linewidth}{@{\extracolsep{\fill}}lrrrr@{}}
    \toprule
    & \multicolumn{2}{c}{\textbf{Skill2Real (ours)}}
      & ASPIRE & Zetta \\
    \cmidrule(lr){2-3}
    Task & GPT-6 Astra & Claude Opus 4.6 & Claude Opus 4.6 & $\pi_{0.5}$ \\
    \midrule
    Stove + moka pot      & \textbf{100.0} & \textbf{100.0} & \textbf{100.0} & 95.0 \\
    Bowl in drawer        &  20.0 &   0.0 &   0.0 & \textbf{75.0} \\
    Mug in microwave      &  \textbf{15.0} &   0.0 &   0.0 &  5.0 \\
    Both mokas on stove   &   \textbf{5.0} &   0.0 &   0.0 &  0.0 \\
    Soup + cream cheese   &   \textbf{0.0} &   \textbf{0.0} &   \textbf{0.0} &  \textbf{0.0} \\
    Soup + tomato sauce   &  80.0 &   0.0 &   0.0 & \textbf{90.0} \\
    Cream cheese + butter &  25.0 &  85.0 &  \textbf{94.0} & 50.0 \\
    Mugs on two plates    &  \textbf{35.0} &  10.0 &  32.0 & 15.0 \\
    Mug + pudding         &  \textbf{25.0} &  15.0 &   0.0 &  5.0 \\
    Book in caddy         &  50.0 &  30.0 &   0.0 & \textbf{65.0} \\
    \midrule
    Mean                  &  35.5 &  24.0 &  22.6 & \textbf{40.0} \\
    \bottomrule
  \end{tabular*}
\end{table}

\begin{table}[!htbp]
  \caption{\textbf{Per-task LIBERO-Pro Long success under instruction
  perturbations (Task).} Success rates (\%) for the fixed C3/B3 comparison.
  Skill2Real uses 20 trials
  per task. Baselines: ASPIRE ($N{=}90$)~\citep{lu2026aspire};
  Zetta (T)~\citep{ding2026zetta}.
  $\pi_{0.5}$ is Zetta's base policy.}
  \label{tab:long_per_task_task}
  \centering\small
  \setlength{\tabcolsep}{4pt}
  \renewcommand{\arraystretch}{1.1}
  \begin{tabular*}{\linewidth}{@{\extracolsep{\fill}}lrrrr@{}}
    \toprule
    & \multicolumn{2}{c}{\textbf{Skill2Real (ours)}}
      & ASPIRE & Zetta \\
    \cmidrule(lr){2-3}
    Task & GPT-6 Astra & Claude Opus 4.6 & Claude Opus 4.6 & $\pi_{0.5}$ \\
    \midrule
    Stove + moka pot      & \textbf{100.0} & \textbf{100.0} &  26.0 & \textbf{100.0} \\
    Bowl in drawer        &  \textbf{55.0} &   0.0 &   0.0 &   0.0 \\
    Mug in microwave      &  \textbf{15.0} &   0.0 &   0.0 &   0.0 \\
    Both mokas on stove   & \textbf{100.0} & \textbf{100.0} & \textbf{100.0} & \textbf{100.0} \\
    Soup + cream cheese   &  \textbf{95.0} &  50.0 &  70.0 &  80.0 \\
    Soup + tomato sauce   &  30.0 &  50.0 &  \textbf{70.0} &  35.0 \\
    Cream cheese + butter &  85.0 & \textbf{100.0} &  86.0 &  95.0 \\
    Mugs on two plates    &  \textbf{90.0} &  15.0 &   0.0 &  25.0 \\
    Mug + pudding         &  90.0 &  20.0 &   4.0 &  \textbf{95.0} \\
    Book in caddy         &  90.0 &  30.0 &  26.0 & \textbf{100.0} \\
    \midrule
    Mean                  &  \textbf{75.0} &  46.5 & 38.3 & 63.0 \\
    \bottomrule
  \end{tabular*}
\end{table}

\subsection{Real-Robot Evaluation Details}
\label{appendix:real_protocol}

\paragraph{Hardware and observation interface.}
All real-world tasks use a single UR5e arm equipped with a Pika gripper.
The robot executes joint-space trajectories through ROS Noetic and the UR
trajectory controller. Visual observations come from a fixed overhead
Intel RealSense D435 and a wrist-mounted RealSense D405; the wrist camera
provides \mbox{$640\times480$} images. Both views provide RGB images and
aligned metric depth from the RealSense sensors. Hand--eye calibration
relates the wrist camera to the robot's end effector. As in LIBERO, the
public observation interface exposes both scene and wrist RGB-D views.
Robot-specific calibration and low-level control are part of the backend
implementation.

\paragraph{Tasks and deployment.}
We evaluate real-robot transfer on specified-object pick-and-place (R1),
attribute-based sorting (R2), equation assembly with cubes (R3), and drawer
manipulation (R4).  Appendix~\ref{appendix:real_world_tasks} defines their
instructions and completion conditions.  Both memories are learned in
simulation and frozen for real evaluation. Deployment uses no real-world
task demonstrations, task-policy fine-tuning, or online skill-memory
updates. The real backend implements the
same documented API, and the Proposer uses camera observations and public
API returns without a real-world Verifier, Governor, or skill update.

Table~\ref{tab:real_results} compares CaP-Agent0 from CaP-X~\citep{fu2026capx},
a GPT-6 Astra Proposer without learned skills, the two single-memory
conditions, and the full hierarchy.  All Astra conditions share the test-time
model and robot interface.  Cerebellum-only evaluation uses the Stage-I
memory.  Brain-only training runs Stage II with empty cross-level context,
so its memory does not depend on unavailable cerebellum skills. Each method
has 20 trials on each of R1--R4. A completed
trial must satisfy its task-specific terminal condition, and Mean averages
the four task success rates without weighting.

\paragraph{Qualitative image selection.}
Figure~\ref{fig:sim_real_skills} pairs recorded LIBERO-90 and real-robot
frames by operation type. The top row of Figure~\ref{fig:real_execution}
shows a complete citrus-retrieval task, from grasp in the open drawer to
release on the table. The bottom row shows the initial equation
and two answer-cube placements for R3, with completion confirmed by an
archived final-state review. These images illustrate selected execution
stages rather than the aggregate evaluation in Table~\ref{tab:real_results}.

\Needspace{6\baselineskip}
\subsection{PVG Training Ablations and Transfer Evaluation}
\label{appendix:pvg_protocol}

All variants train with GPT-5.6 Sol on LIBERO-90 and share the API and
source task stream. Without the Verifier, the Proposer consolidates from its public
rollout record without privileged diagnosis.  Without the Governor, updates
pass deterministic skill-store checks but bypass evidence-based admission.
The evaluator and skill store remain present.  A public-observation-only
Verifier would be needed to isolate privilege from the Verifier's other
contributions.

\paragraph{Transfer to LIBERO-Pro Long.}
Figure~\ref{fig:pvg_pro_long_curve} evaluates each frozen checkpoint with
GPT-6 Astra on LIBERO-Pro Long. The removed roles affect only LIBERO-90
training; neither the Verifier nor the Governor participates in target
evaluation.
All variants begin at 2.0\% Overall; at B4, Full Skill2Real achieves
56.25\%, compared with 39.0\% without the Verifier and 43.0\% without
the Governor. Losses are calculated from unrounded values, yielding
17.25 and 13.25 percentage points; the main text reports one decimal place.

\begin{figure}[!htbp]
  \centering
  \includegraphics[width=\linewidth]{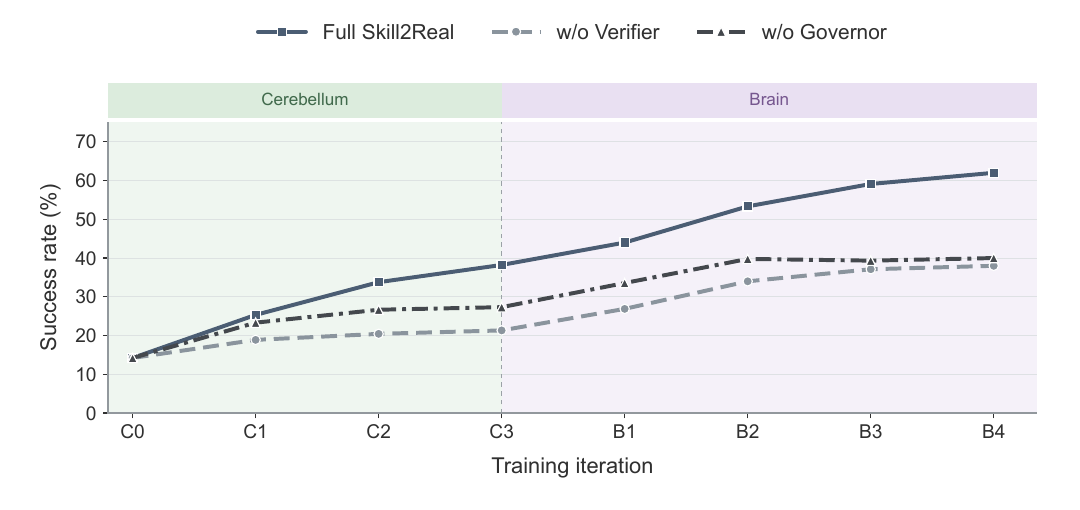}
  \caption{\textbf{PVG learning dynamics on LIBERO-90 with GPT-5.6 Sol.}
  Success is normalized by 450. Background shading separates cerebellum
  and brain learning at C3; colors, markers, and line styles distinguish
  the three training conditions.}
  \label{fig:pvg_learning_curve}
\end{figure}

\paragraph{Source-suite LIBERO-90 evaluation.}
Figure~\ref{fig:pvg_learning_curve} evaluates the learned libraries on
LIBERO-90 using Sol, reporting the C0 baseline and all seven training
iterations. Success is the total score divided by 450. All three variants start from the same C0 score of 64/450
(14.2\%). C1--C3 track cerebellum learning, followed by brain learning at
B1--B4. Each evaluation uses five initial states per task, as specified in
Appendix~\ref{appendix:libero_protocol}. All variants are compared at B4,
the final brain iteration of the same two-stage schedule.

Full Skill2Real has the highest success after every training iteration.
Without the Governor, success stays near 40\% from B2 through B4, while Full
Skill2Real increases from 53.3\% to 62.0\%. Full Skill2Real reaches 62.0\%, versus 38.0\% without the Verifier
and 40.0\% without the Governor. The 24.0- and 22.0-point final
gaps show lower success when either role is removed in these recorded
runs. The curves do not measure variability across independent training
runs.

\section{Shared Robot API and Execution Contract}
\label{appendix:standard_api}
\setcounter{table}{0}

The shared API in Section~\ref{subsection:agentic_policy} separates the
Proposer's manipulation decisions from backend-specific perception and
control. This appendix describes the 18 operations in the supplied
\texttt{standard\_api/v1} simulation contract and the information passed
between them. The operation inventory instantiates $\mathcal A$; its
documentation supplies $D_{\mathcal A}$. Cerebellum and brain skills guide
how the Proposer composes these operations, without extending the callable
interface. The descriptions below summarize operation semantics rather
than reproduce complete signatures or backend implementations. All edited
examples in the appendices use this documented operation inventory.

\subsection{Public Perception and Measured Geometry}
\label{appendix:api_perception}

Perception operations expose observations and measurements from which the
Proposer selects task entities and action parameters. In the simulation
contract, metric depth is reconstructed from calibrated stereo RGB images;
the simulator depth buffer is not a public input. This specifies how depth
is obtained, while RGB-D in the experimental setup describes the resulting
observation modality. The real backend uses calibrated RGB-D sensors
(Appendix~\ref{appendix:real_protocol}); sharing API semantics does not
require identical sensing implementations.

\begin{table}[!htbp]
  \caption{\textbf{Public perception operations (8 of 18).}
  Image retrieval and geometric measurement supply evidence for the
  Proposer's decisions. They do not certify manipulation outcomes.}
  \label{tab:standard_api_perception}
  \centering\small
  \setlength{\tabcolsep}{4pt}
  \renewcommand{\arraystretch}{1.14}
  \begin{tabular}{@{}>{\raggedright\arraybackslash}p{0.41\linewidth}
      >{\raggedright\arraybackslash}p{0.56\linewidth}@{}}
    \toprule
    Operation & Public result and interpretation \\
    \midrule
    \texttt{get\_scene\_image}
      & Path to the current front scene image. \\
    \texttt{get\_topdown\_image}
      & Path to the current top-down image; target pixels are selected
        in an observation frame. \\
    \texttt{get\_wrist\_view}
      & Wrist image for the selected arm, with available depth preview
        and camera calibration. \\
    \addlinespace[3pt]
    \texttt{get\_topdown\_stereo\_view}
      & Stereo RGB pair with intrinsics and extrinsics for both cameras. \\
    \texttt{get\_foundation\_stereo\_depth}
      & Metric depth computed from a calibrated image pair, with
        \texttt{observation\_ref} and \texttt{depth\_ref} identifying
        the source observation and its depth/calibration binding.
        Missing calibration causes failure. \\
    \addlinespace[3pt]
    \texttt{get\_sam\_mask\_from\_box}
      & Segmentation mask for a supplied image and bounding box. \\
    \texttt{measure\_visible\_geometry}
      & Visible 3-D center, world-frame bounds, top height, principal
        axes and extents, valid-point count, and confidence from a mask,
        depth, and calibration. It supplies measurements, not a grasp
        choice or action plan. \\
    \texttt{get\_topdown\_object\_candidates}
      & Candidate observations for source discovery, with geometry
        and validity checks. This is neither a complete object census
        nor evidence that a target placement succeeded. \\
    \bottomrule
  \end{tabular}
\end{table}

The public interface excludes simulator object handles, ground-truth object
poses, hidden scene structure, and task-success predicates. Visible
candidates therefore remain hypotheses to bind from observations, rather
than privileged object identities. Likewise, a principal axis or a valid
point cloud describes measured geometry; the Proposer must still choose
the contact, closing axis, destination, and intended effect.
For a box-based segmentation query, the Proposer selects the box from the
public image. Multiple candidate regions require separate supplied boxes;
the contract does not promise text-prompted mask enumeration or a
segmentation score. Geometry confidence describes the measurement returned
by \texttt{measure\_visible\_geometry}, not semantic target correctness.

\Needspace{8\baselineskip}
\subsection{Execution State, Pose Construction, and Control}
\label{appendix:api_execution}

State and pose operations connect public observations to executable commands.
The runtime associates calls with the current execution step, so programs do
not supply an internal step index. The two image-path operations in
Table~\ref{tab:standard_api_perception} return strings; the remaining
operations return structured results with status information. A returned
status must be interpreted at the level of the operation that produced it.

\begin{table}[!htbp]
  \caption{\textbf{State, compilation, and control operations (10 of 18).}
  Pose construction does not move the robot. The final four operations
  issue arm or gripper commands.}
  \label{tab:standard_api_execution}
  \centering\small
  \setlength{\tabcolsep}{4pt}
  \renewcommand{\arraystretch}{1.14}
  \begin{tabular}{@{}>{\raggedright\arraybackslash}p{0.41\linewidth}
      >{\raggedright\arraybackslash}p{0.56\linewidth}@{}}
    \toprule
    Operation & Public result or commanded effect \\
    \midrule
    \texttt{preprocess}
      & Refresh public observations used by perception operations. \\
    \texttt{get\_observation\_geometry}
      & Observation-frame geometry for a supplied public description,
        providing an alternative source-localization path. \\
    \texttt{get\_previous\_proposer\_state}
      & Previous decision-stage public state and the latest still-valid
        candidate scan. After an executed compiled grasp, while the
        gripper remains closed, this can include
        \texttt{held\_object\_state} for place compilation. \\
    \texttt{grasp\_center\_to\_tool0\_pose}
      & Tool-frame pose corresponding to a selected grasp center
        and orientation. \\
    \texttt{compile\_pose}
      & In grasp mode, construct poses from Proposer-supplied
        \texttt{grasp\_center\_xyz} and
        \texttt{closing\_axis\_world}. In place mode, construct
        carry, release, and retreat poses from a target pixel,
        bound depth, and held-object geometry. \\
    \texttt{get\_end\_effector\_pose}
      & Current end-effector pose for the selected arm. \\
    \addlinespace[3pt]
    \texttt{gohome}
      & Command an arm to its home configuration. \\
    \texttt{goto\_pose}
      & Command an arm to a position and orientation, with documented
        approach and contact options. Position arguments are grounded
        in API results. \\
    \texttt{gripper\_goto}
      & Command a gripper opening in meters, bounded by the supported
        range. A zero command requests closure. \\
    \texttt{goto\_pose\_both}
      & Command both arms to their respective poses in one call. \\
    \bottomrule
  \end{tabular}
\end{table}

Execution state is distinct from learned skill memory. Here, ``previous
stage'' means an earlier decision stage $t$, not Stage~I or Stage~II of
training. The runtime's held-object state carries geometric information
needed to construct placement poses; it is not a sensor assertion that the
object remains held. Refreshing this state or rebinding a target during
deployment does not modify $S_{\mathrm{cer}}^*$ or $S_{\mathrm{brain}}^*$.
The placement compiler consumes the runtime's carry geometry; a skill's
explanatory contact offset is not a replacement for that state. Motion
quaternions use the documented \texttt{xyzw} order.

\Needspace{8\baselineskip}
\subsection{From an Observed Target to an Executed Placement}
\label{appendix:api_composition}

Placement combines a policy decision, geometric construction, and physical
execution. The following sequence illustrates their dependencies under the
contract; it is not an additional API operation or a recorded rollout.
\begin{enumerate}[leftmargin=*,itemsep=5pt,topsep=5pt]
  \raggedright
  \item \textbf{Acquire a calibrated observation.}
    \texttt{get\_topdown\_stereo\_view} supplies an RGB pair, and
    \texttt{get\_foundation\_stereo\_depth} produces metric depth with
    its \texttt{observation\_ref}/\texttt{depth\_ref} binding.
  \item \textbf{Choose and bind the target.}
    The Proposer selects the destination and target pixel from the
    corresponding public frame, conditioned on the task and skill memories.
    It supplies this binding and the runtime's carry geometry to
    \texttt{compile\_pose} in place mode. The compiler uses them to
    determine support-relative release and clearance poses.
  \item \textbf{Execute the constructed poses.}
    After inspecting compilation status, the program composes
    \texttt{goto\_pose} and \texttt{gripper\_goto} calls for transport,
    release, and retreat, checking each returned result.
  \item \textbf{Observe the physical result.}
    Check public retention evidence during loaded motion and inspect
    fresh observations after release for the requested placement relation.
    A completed command alone does not establish either outcome.
\end{enumerate}

The observation and depth references bind a chosen pixel to its measured
frame and calibration; programs do not construct or edit that binding.
These consistency checks prevent mixing incompatible inputs, but they do
not establish that a previously observed object has remained stationary.
The Proposer must obtain fresh evidence when scene changes invalidate a
binding. Similarly, successful compilation establishes that poses were
constructed under the contract, not that execution achieved the task.

\subsection{Relation to PVG Learning and Frozen Deployment}
\label{appendix:api_pvg_boundary}

The API defines the Proposer's information and action channel, while PVG
defines how execution evidence changes persistent knowledge. In
Algorithm~\ref{alg:skill2real}, the deterministic evaluator can inspect
privileged simulation state to produce $e$. The Verifier converts that
evidence into $f\in\mathcal F_{\mathrm{pub}}$, expressed through public
observations and API semantics. Thus, excluding a success oracle from the
robot API does not remove privileged evaluation from simulation training.
Neither the evaluator nor the Verifier is a callable robot operation.

The Proposer uses the feedback to author candidate update $\delta$.
Subsequent validation produces $e_{\delta}$, on which the Governor bases
admission. API input checks and pose-compilation status are not substitutes
for this validation, and candidate admission is distinct from the
stage-level stopping criterion. The same three roles learn cerebellum
skills in Stage~I and brain skills in Stage~II, with the cerebellum fixed
during Stage~II. At deployment, the Proposer executes through the real
backend with both memories frozen and without the Verifier or Governor.

The supplied interface specification also records checks on the operation
inventory, dispatch surface, documentation, and implementation digests.
Such checks track a particular contract version; they do not establish
physical task success or equivalence of perception and control across
backends. The transfer requirement in Section~\ref{subsection:deployment}
is compatible public semantics, with calibration and low-level execution
implemented for the target robot.

\FloatBarrier
\section{Recorded PVG Learning and Admission}
\label{appendix:pvg_worked_case}
\setcounter{table}{0}

This recorded bowl-transfer case illustrates how execution feedback informs
program revision and how the Governor selects knowledge for memory.
The account condenses supplied programs, feedback, candidate entries, and
admission decisions into a qualitative example.\footnote{Source:
\href{https://sapporo.xincheng2004.com/appendix/02_appendix_worked_examples.html}
{complete worked examples}, Sections A.1--A.8, task 60. This separate
basket/tray development run used frozen initial Cerebellum templates;
its records are preserved in the source archive.}

\subsection{From Execution Failure to a Candidate Skill}
\label{appendix:pvg_task60_trace}

\paragraph{Failure and diagnosis.}
The instruction asks the robot to place the left black bowl in a tray.
Both visible bowls receive similar segmentation confidence, so the Proposer
uses their current image positions to select the intended instance.
Early attempts identify the correct bowl but fail to retain it during
lifting. The Verifier localizes the failed transition to contact and
retention: touching the bowl or completing a gripper-close command does
not establish a grasp. It also notes that interpreting \emph{left} depends
on the current viewpoint.

\paragraph{Revision and candidate submission.}
The Proposer revises contact construction toward the bowl's narrower lower
profile and checks the object after a short lift. Subsequent executions
provide successful task outcomes and evidence for candidate review.
The Proposer then submits entries about target selection, contact binding,
destination geometry, and progress checking. The Governor evaluates each
entry's content against the actual program and execution record.

\subsection{Why One Candidate Was Admitted and Three Were Rejected}
\label{appendix:pvg_task60_admission}

The four proposals share a successful source program, but differ in their
abstraction level and evidential support (Table~\ref{tab:pvg_task60_decisions}).

\begin{table}[H]
  \caption{\textbf{Governor decisions in the recorded bowl-transfer case.}
  Descriptions condense the supplied candidate and decision records.}
  \label{tab:pvg_task60_decisions}
  \centering\small
  \setlength{\tabcolsep}{4pt}
  \renewcommand{\arraystretch}{1.2}
  \begin{tabular}{@{}>{\raggedright\arraybackslash}p{0.25\linewidth}
    >{\raggedright\arraybackslash}p{0.12\linewidth}
    >{\raggedright\arraybackslash}p{0.57\linewidth}@{}}
    \toprule
    Candidate & Decision & Recorded basis \\
    \midrule
    Relative referent & Admit & Selects the requested left bowl from
      fresh image evidence. The executed program implements the rule,
      and the entry states when the ordering must be reconsidered. \\
    Contact binding & Reject & The proposed contact-point and
      local-geometry rule belongs to the frozen Cerebellum rather
      than Brain memory. \\
    Destination geometry & Reject & Computing tray geometry and
      release clearance is a local placement procedure beyond the
      permitted Brain update. \\
    Progress checking & Reject & The entry includes an object--hand
      co-motion test that was not implemented or tested. \\
    \bottomrule
  \end{tabular}
\end{table}

\paragraph{What enters memory.}
The admitted Brain entry resolves the relative referent: merge duplicate
bowl masks and select the distinct bowl with the leftmost center in the
current image. It calls for reconsideration when viewpoint, visibility,
or occlusion changes that ordering. The local contact revision remains
outside the admitted Brain entry, and the untested progress-checking
clause is not promoted.

\paragraph{What the case illustrates.}
The Verifier diagnoses a failed physical transition, the Proposer authors
the revision and proposed knowledge, and the Governor checks both layer
boundaries and agreement with execution evidence. Whole-program success
supports review but does not validate every associated proposal or isolate
the causal contribution of one admitted rule.

\FloatBarrier
\section{PVG Skill-Memory Templates and Examples}
\label{appendix:pvg_skill_examples}
\setcounter{table}{0}

This appendix expands the skill representation in
Section~\ref{subsection:hierarchical_training} through an initial brain
template and edited cerebellum and brain examples. Every API call in these
examples belongs to the final 18-operation interface in
Appendix~\ref{appendix:standard_api}. The examples illustrate skill content;
they are not additional evaluated rollouts. The initial template is based on
the preserved \href{https://sapporo.xincheng2004.com/sources/cdf1e36f0d02644d/task_policy.md}{source snapshot}.
The PVG roles and admission procedure follow Algorithm~\ref{alg:skill2real}.

\subsection{Initial Brain Template}
\label{appendix:pvg_initial_template}

The initial \texttt{task\_policy.md} supplies a writing scaffold with no
learned task-level entries. Thus, an empty brain memory means an absence of
learned task knowledge, not an absence of API documentation or a generic
instruction scaffold. During Stage~II, the Proposer develops task-level
composition while the Stage-I cerebellum remains frozen.

\begin{pvgskill}{Initial memory / Brain / condensed template}
Task: <public task instruction>
Learned task-level entries: none
Fixed context in Stage II: frozen cerebellum skills; documented robot API

## Task Understanding
Describe the required outcome and the public evidence that distinguishes
completion from partial progress. Bind objects and destinations from fresh
observations. Separate observed facts, assumptions, and unresolved questions.
Do not retain hidden scene facts or initial-state-specific coordinates.

## Starting Procedure
Read the public goal and observe the scene.
Determine the remaining objectives and the uncertainties that affect them.
Choose the next objective and an applicable frozen cerebellum skill.
Bind public inputs; specify the intended effect and its evidence check.
Execute through the documented API and inspect the returned evidence.
Continue on supported progress; request observations when a distinction is
missing; revise the remaining plan after an unsupported transition.
Preserve achieved effects and reason from the actual post-action scene.
Report completion only when supported by evidence under the task protocol.

## Learned Task-Level Knowledge
<Initially empty.>
For each proposed method, describe the public situation, the change in task
decisions, evidence for continuing, and failure or untested conditions.
Organize supported knowledge as procedures, branches, or short sketches.
The layout is editable; no fixed number or taxonomy of skills is required.

## Open Questions And PVG Update
<Initially empty.>
Use filtered Verifier feedback to identify an unresolved decision and a
discriminating trial. The Proposer authors the program and candidate update.
The Governor decides admission from validation evidence; an untested guess
or a successful episode alone does not establish a reusable improvement.

## Hierarchy Boundary
The brain selects objectives, their order, skills, and public input bindings.
Local geometry, contact, motion, retention checks, and local recovery belong
to the cerebellum. Do not modify that library during brain training.
Record a missing capability and its evidence when it limits composition.
\end{pvgskill}

\Needspace{8\baselineskip}
\subsection{Illustrative Skill Representation under the Shared API}
\label{appendix:pvg_skill_representation}

Section~\ref{subsection:hierarchical_training} represents a skill through
natural-language guidance, an API-level call sequence, and a compact state
description. The edited example below uses the final shared API to illustrate
this format. It has no associated rollout evaluation and is not used as
evidence for the reported results. Its compact fields describe episode state
rather than an API return schema.

\begin{pvgskill}{Illustrative API template / clear-side rim acquisition}
## Scope
Use when the public scene supports the rim-contact trigger. The task objective
and destination are provided by the Proposer's task-level composition.

## Natural-Language Guidance
Use measured visible body/rim geometry and a visibly clear side. Engage the wall
below the rim, with explicit image-selected closing and downward axes.
Inspect object--hand contact after closure and co-motion after a short lift.
Use runtime carry geometry when the next operation compiles a placement.

## Public Inputs
Object/handle images; local clearance; mask-based measured geometry;
fresh post-action images.
Missing geometry, calibration, or visual evidence remains unresolved.

## Compact State (Explanatory Fields)
contact: selected rim side, contact position, closing axis, approach axis
phase: contact selected / closed / lifted
retention: unresolved / supported / contradicted
evidence: public observations and API returns supporting the current phase
carry: runtime-owned held_object_state when available; never synthesized

## Decision And Execution Boundary
1. Read a public image; select the object box and a clear rim side.
2. Use get_sam_mask_from_box and measure_visible_geometry with the calibrated
   depth belonging to that observation. Inspect the returned measurements.
3. Choose grasp_center_xyz and closing_axis_world from this public evidence.
   Supply the documented request to compile_pose in grasp mode.
4. After checking compilation, use goto_pose for the constructed approach
   and contact; inspect each result before issuing a dependent command.
5. Use gripper_goto for closure. Reobserve contact, then execute the short
   lift with goto_pose and inspect fresh images for continued retention.
6. If retention is unresolved, obtain another public view. If contradicted,
   revise local recovery. Otherwise return progress for task-level reasoning.

## Expected Effect And Limits
Supported acquisition means the vessel remains retained after the lift.
A constructed pose or completed command alone does not establish that effect.
An unsupported local effect calls for observation or local recovery, without
claiming task completion or changing the frozen skill text during deployment.
\end{pvgskill}

\paragraph{Memory and execution state.}
In this illustration, the contact rule is persistent guidance. Contact coordinates, selected
objects, measured offsets, and current retention evidence are episode-specific
bindings refreshed from public observations. Updating those bindings during
execution does not update $S_{\mathrm{cer}}^*$ or $S_{\mathrm{brain}}^*$.
This distinction is essential to the frozen deployment policy in
Section~\ref{subsection:deployment}.

\paragraph{What the representation establishes.}
The illustrative template specifies when to act, how to construct a local
action, and what evidence to inspect afterwards. Its compact state supports
composition with later skills. The representation alone does not supply a
new success metric or prove that the rule works outside its supported scope.

\Needspace{8\baselineskip}
\subsection{Documented Calls and the Placement Handoff}
\label{appendix:pvg_api_call_template}

The sequence below names only operations in the final 18-operation inventory.
The Proposer supplies arguments from the bound public observation and inspects
each result before a dependent call.

\begin{pvgskill}{Shared-API call sequence / observe, grasp, and place}
1. get_topdown_stereo_view; get_foundation_stereo_depth.
2. get_sam_mask_from_box for each selected region;
   measure_visible_geometry for a supported mask.
3. compile_pose in grasp mode; goto_pose for approach and contact;
   gripper_goto for closure; reobserve retention after a short lift.
4. get_previous_proposer_state for runtime carry geometry;
   compile_pose in place mode for the selected destination.
5. goto_pose for carry, release, and retreat; gripper_goto for release;
   reobserve the placement relation.
\end{pvgskill}

\begingroup\raggedright
\paragraph{Ground the inputs.}
Metric depth and calibration come from
\texttt{get\_topdown\_stereo\_view} followed by
\texttt{get\_foundation\_stereo\_depth}. The Proposer chooses
\texttt{box\_xyxy} on the source image identified by that depth result;
the mask and measurements must refer to this bound observation. A separately
retrieved top-down image can check the physical result but cannot replace the
bound source frame. The measurement options are only those documented
for the backend. For grasp construction, the request contains the chosen
\texttt{grasp\_center\_xyz} and \texttt{closing\_axis\_world}.
Motion positions come from public measurements or constructed poses, and
quaternions use \texttt{xyzw} order.

\paragraph{Inspect before continuing.}
The Proposer groups native mask results under caller-supplied region labels;
these are episode bindings, not verified identities. Using $D_{\mathcal A}$,
it checks status and extracts a supported mask before requesting geometry
with matching depth and calibration.
Failed compilation blocks motion; completed closure calls for fresh visual
retention checks.

\paragraph{Pass the placement binding.}
After an executed compiled grasp, the previous public state may contain
\texttt{held\_object\_state}. The Proposer passes that runtime-owned
geometry, the selected destination pixel, and its immutable
\texttt{observation\_ref}/\texttt{depth\_ref} binding in the documented
place request. The compiler constructs carry, release, and retreat poses.
Missing carry geometry or an invalid binding leaves placement unresolved;
the program does not invent replacement state. Execution uses the returned
poses, and fresh post-release observations check the intended relation.
\par\endgroup

\Needspace{8\baselineskip}
\subsection{Cerebellum Skills Beyond Grasping}
\label{appendix:pvg_non_grasp_skills}

The following cerebellum examples address two local decisions: identifying
a receptacle's actionable opening and selecting an interior placement target.
Their decision rules are recorded in entry \texttt{chunk47-task52-pattern0} of
\href{https://sapporo.xincheng2004.com/sources/3ac772c43791971a/inspect.md}{\texttt{inspect.md}}
and entry \texttt{chunk35-task48-pattern1} of
\href{https://sapporo.xincheng2004.com/sources/25e2e1e7bc06b0d4/transport.md}{\texttt{transport.md}}.
The API sequence separates region segmentation from geometric
measurement: named boxes supply candidate masks through
\texttt{get\_sam\_mask\_from\_box}, and the Proposer inspects those results
before calling \texttt{measure\_visible\_geometry} for a selected region
(Appendix~\ref{appendix:pvg_api_call_template}). Task-entity selection and
the intended relation remain brain-level decisions.

\begin{pvgskill}{Local perception / geometry-supported receptacle opening}
Source: chunk47-task52-pattern0; memory: cerebellum / inspect

## Public Trigger
The parent receptacle is visible, but an edge or wall patch can be mistaken
for its broad actionable opening. The intended region needs geometric checks.

## Learned Rule
Mark the parent and candidate opening regions on the bound public image.
Call get_sam_mask_from_box for each named box and inspect each returned mask.
Require sufficient region area and parent-relative overlap or containment.
Keep the distinction unresolved when the visible evidence is ambiguous.

## Procedure And State
Choose a supported opening from the inspected region results. Extract its
mask artifact according to the API schema, then call measure_visible_geometry with
the depth and calibration belonging to the same observation. Inspect the
native geometry result before using it. Carry the region label, selected
mask, parent relation, and observation/depth binding as episode state.

## Expected Effect And Boundary
Return geometry for the actionable opening, rather than an unsupported sliver.
If no candidate has adequate public support, keep localization unresolved;
no text-prompted mask enumeration or segmentation-score field is assumed.
\end{pvgskill}

\begin{pvgskill}{Transport and release / place relative to the receptacle interior}
Source: chunk35-task48-pattern1; memory: cerebellum / transport

## Public Trigger
An open receptacle is viewed obliquely. A visible wall or an incomplete view
of the opening makes its apparent center an unreliable placement target.

## Learned Rule
Compare the observed receptacle bounds with the visible opening to choose
a supported interior target pixel. Reobserve if occlusion prevents that
choice. Preserve the source rule's support-relative placement intent through
compile_pose, which owns support height, carry geometry, and release poses.

## Procedure And State
Bind the target pixel to its observation_ref and depth_ref. Read the runtime's
held_object_state through get_previous_proposer_state; never fabricate it.
Pass these inputs in the documented compile_pose place request. Inspect the
constructed carry, release, and retreat poses before executing them through
goto_pose and gripper_goto. Check each result before the next dependent move.

## Expected Effect And Boundary
The released object settles in the intended interior. Reobserve this relation;
a completed descent or opening command alone does not establish containment.
\end{pvgskill}

\Needspace{8\baselineskip}
\subsection{A Final Brain Skill for Relational Composition}
\label{appendix:pvg_final_brain}

The task-63 Brain document in the completed third round of the stacking
training group contains two learned methods:
\texttt{chunk1-task63-brain0} and \texttt{chunk1-task63-brain1}.
The \href{https://sapporo.xincheng2004.com/sources/425e7c513364c577/skill.md}{source document}
records task-level ordering and referent binding while using the frozen
cerebellum. The example below combines those two methods for presentation;
it does not replace them with a new experimentally evaluated program.

\begin{pvgskill}{Final Brain example / compose stacking and tray placement}
Sources: chunk1-task63-brain0 (sequence), chunk1-task63-brain1 (binding)
Memory: brain; context: frozen cerebellum

## Task And Public Trigger
The requested final relations are the designated left bowl on the designated
right/support bowl, with the pair in the tray. Public evidence provides no
verified way to carry the nested pair together.

## Learned Task-Level Order
When the instruction allows this order, place the support bowl in the tray
first. Verify and preserve that relation, then execute only the remaining
upper-bowl-on-support part of the task.

## Learned Rebinding Rule
After moving the support bowl, inspect a fresh tray image. Select boxes for
the visible bowls, obtain their masks with get_sam_mask_from_box, and measure
them with matching depth and calibration. Bind the support only when current
image and geometry evidence distinguish the bowl inside the tray. An old
relational description can instead refer to the untouched bowl; measurement
confidence alone neither resolves identity nor proves placement success.

## Compact Task State
Goal relations: support in tray; upper bowl on support.
Progress: only relations supported by current public evidence.
Bindings: upper bowl, tray, and support bowl rebound after its movement.
Remaining objective: the unsatisfied relation, with prior progress preserved.

## Composition Sketch
Bind the two task objects and tray from public observations.
Use applicable frozen cerebellum skills to place the support bowl in the tray.
Observe the resulting relation and rebind the support by current containment.
If exactly one support is grounded, compose the upper-bowl placement suffix.
Reobserve both requested relations before reporting completion.

## Applicability Boundary
Do not apply this reordering when the instruction requires stacking first.
If the support is outside the tray, has moved, or cannot be identified
unambiguously, withhold the dependent suffix and preserve the observed state.
Initially occupied trays and other object classes were not established by
the recorded evidence for these methods.
\end{pvgskill}

\paragraph{Why this knowledge is brain-level.}
The learned decisions concern which relation to establish first and how to
identify an entity after its role in the scene changes. They do not change
grasp geometry, gripper control, or local placement code. This is the Stage-II
dependency in Equation~\ref{eq:hierarchical_training}: task composition is
learned with $S_{\mathrm{cer}}^*$ fixed.

The source reports whole-program development evidence and explicitly notes
that it is not a matched causal evaluation of either decision alone. We use
the record to illustrate learned content, without converting those annotations
into a new held-out score or an isolated effect estimate.

\Needspace{8\baselineskip}
\subsection{Task Progress and Local Recovery in Drawer Manipulation}
\label{appendix:pvg_drawer_boundaries}

Drawer manipulation illustrates how the two memory levels handle different
decisions. The Brain entry below is retained in the task-5 document from a
completed third round of articulated-storage training. The cerebellum entry
comes from the final \texttt{manipulate.md} library. Their juxtaposition
explains the division of responsibility; it is not a claim that this exact
pair was jointly executed in a recorded rollout.

\begin{pvgskill}{Brain / establish placement before allowing drawer closure}
Source: chunk9-task5-brain2; memory: brain / sequence

## Public Trigger And Decision
The requested carton has been released near the bound open drawer, and
closure remains unexecuted. Preserve the place-before-close order.
Allow closure only when a fresh observation supports the carton above the
measured drawer floor and within the originally observed drawer footprint.

## State And Boundary
Carry the intended drawer identity, its observed footprint, and evidence for
the released object's relation to it. If the carton is outside or below the
drawer, or the destination is unresolved, withhold closure and preserve state.
A post-close handle detection that jumps to a different cabinet does not
justify another push. Do not treat a completed release as proved placement.
\end{pvgskill}

\begin{pvgskill}{Cerebellum / recover a partially completed slider interaction}
Source: chunk51-task0-pattern0; memory: cerebellum / manipulate

## Public Trigger And Local Policy
A drawer or slider initially moves, then deflects or stalls despite later
commands completing. Preserve the continuous, useful part of the stroke.
Disengage without an outward motion that reverses progress, reobserve the
shifted part, and relocalize it using relative geometry and continuity.
Use fresh short monotonic pusher strokes from the achieved state.

## State And Effect Check
Carry the observed part position, achieved displacement, contact mode, and
current continuity evidence. Check a compact pusher by observed part
displacement; a completed gripper command alone does not establish motion.

## Failure Boundary
Withhold the next interaction if relocalization merges adjacent handles.
Command completion alone does not establish slider motion or final closure.
\end{pvgskill}

The \href{https://sapporo.xincheng2004.com/sources/6523fad4341dad52/skill.md}{Brain source}
conditions a task transition on placement evidence; the
\href{https://sapporo.xincheng2004.com/sources/a98e77164f368184/manipulate.md}{cerebellum source}
controls the local realization of an already selected interaction. During
Stage II, the Brain may change whether and when to request closure, but it
cannot patch the frozen local controller. During deployment, both policies
stay fixed while current bindings and progress evidence change.

\Needspace{8\baselineskip}
\subsection{Skill Invocation and the PVG Evidence Boundary}
\label{appendix:pvg_skill_invocation}

The Proposer is the acting agent at both hierarchy levels. Brain and
cerebellum skills are memories conditioning its program generation, not
additional agents. The following walkthrough expands
Equation~\ref{eq:agentic_policy} using the relational-composition example.
It shows what information passes between task decisions and local execution;
the numbered stages are an explanatory sequence, not a measured trajectory.

\begin{pvgskill}{Illustrative invocation sequence / frozen two-level memory}
## 1. Public Task And Observation
   Read the requested relations and current public scene. Bind task entities;
   keep unobserved facts unresolved. Both learned memories remain fixed.

## 2. Next Objective From Brain Memory
   Review the ordering skill's public trigger and temporal-order boundary.
   If applicable, select support-bowl placement as the next objective.

## 3. Local Program Using Cerebellum Memory
   Select applicable local perception, grasp, and placement knowledge.
   Bind fresh geometry and generate the next API-level program using D_A.
   Execute it and inspect returned motion, retention, and placement evidence.

## 4. Task-Level Progress Update
   Reobserve the tray. Confirm the supported relation and rebind the moved
   support bowl from current containment. Do not replay a completed prefix
   merely because the remaining upper-bowl placement needs revision.

## 5. Remaining Objective
   If the support binding is unambiguous, generate the local placement suffix.
   Otherwise obtain the missing public evidence before a dependent action.
   Report supported progress separately from unresolved final relations.

## Invariant
Only observations, bindings, programs, and episode state change here.
The Verifier and Governor are absent; no learned-memory update is applied.
\end{pvgskill}

\paragraph{How an entry becomes persistent during training.}
Algorithm~\ref{alg:skill2real} distinguishes a rollout from an admitted skill
update. For trace $\tau$, the deterministic evaluator produces evidence $e$;
the Verifier returns feedback $f\in\mathcal F_{\mathrm{pub}}$; the Proposer
authors candidate $\delta$; and subsequent validation produces
$e_{\delta}$. The Governor decides whether $(\delta,e_{\delta})$ supports
an update. Stage~I applies accepted updates to the cerebellum; Stage~II
applies them to the brain while holding the cerebellum fixed.

\paragraph{What the source records establish.}
The retained skill files supply rules, limitations, and development-program
annotations. The paper examples use the current PVG roles and documented
interface. The skill files alone
do not supply a complete candidate-by-candidate transcript of $e$, $f$,
$\delta$, $e_{\delta}$, and the Governor decision.
Appendix~\ref{appendix:pvg_worked_case} separately reconstructs one recorded
Brain-development case from supplied trial evidence and admission records,
with its campaign scope stated explicitly. Neither a final skill text nor
a whole-program success count isolates an individual rule's causal effect.
Source identifiers support traceability; the public task and observations,
rather than those identifiers, determine applicability.

\paragraph{Transfer boundary.}
The transferred knowledge is the frozen policy content expressed through
public observations and API semantics. The real backend supplies compatible
perception, calibration, and low-level control. A scene-dependent contact
pose or changing progress state is an execution binding, not online training.
No simulation-only evidence or learning-time reviewer is required by the
deployed Proposer.

\FloatBarrier
\section{PVG Role Prompt Templates}
\label{appendix:pvg_prompts}

These six edited templates instantiate the three roles in
Algorithm~\ref{alg:skill2real} and Section~\ref{subsection:deployment}, not
verbatim historical runtime prompts. The Proposer performs four kinds of
invocation; the Verifier diagnoses outcomes; the Governor admits fixed
candidates from subsequent validation. Revisions return to the Proposer for
new validation. Historical labels add no roles or admission levels. API
signatures and returns remain defined by $D_{\mathcal A}$; skill content is
illustrated in Appendix~\ref{appendix:pvg_skill_examples}.

\subsection{Proposer: Cerebellum Rollout}
\label{appendix:pvg_proposer_cerebellum}
\label{appendix:pvg_proposer}
\begin{pvgskill}{Proposer / Stage I: local manipulation}
You are the Proposer during Stage I of Skill2Real. Execute a local
manipulation task through the shared code-based policy interface using public information.

## Context
Task: <sampled task from the cerebellum training distribution>
Public history: <observations and returned API results through this step>
API documentation: <operation semantics, arguments, and return contracts>
Current memory: <cerebellum skills; no learned entries at initialization>
Cross-level context: none. No learned brain memory is supplied in Stage I.

## Ground The Next Action
Identify the intended local effect and the public evidence needed to check it.
Bind objects, contacts, and motion arguments from current observations and API
results. Check stored cerebellum guidance against the observed situation.
Obtain missing evidence before deciding; do not reuse an earlier episode's
object pose or hidden facts.

## Generate And Execute
Write the next executable program using the documented API. Respect its
units, argument meanings, permitted values, and return conventions. A stored
code pattern may guide composition; it does not add an undocumented method.
Distinguish constructing an action target from executing the action. Inspect
the actual returned results and subsequent observations before proceeding
as though the intended contact, grasp, displacement, or release occurred.
Do not obtain simulator state, hidden object lists, or evaluation predicates.
Use only documented return fields and fresh public images for retention
checks; a gripper command or cached carry geometry does not prove retention.

## Check The Local Transition
Compare the intended effect with public execution evidence. Preserve effects
that are supported, and identify the first failed or uncertain transition.
An issued command is not evidence that the object moved or remained held.
If a documented API call fails, follow its failure contract. If the physical
effect is uncertain, seek the observation needed to resolve that uncertainty.
Choose any recovery from the new public state rather than assuming that the
original state still holds. Keep partial progress separate from completion.

## Record And Hand Off
Return the program and public state needed for the next step, distinguishing
confirmed effects, unresolved evidence, and API failures. The execution trace,
not your summary, supplies the rollout evidence. After the rollout, the
Verifier provides filtered feedback for the Proposer's separate consolidation.

## Memory Boundary
Exploration does not directly change the persistent cerebellum library.
A candidate remains pending until subsequent validation supports admission
by the Governor. Do not update brain memory or claim that an exploratory
success has already established a reusable skill.
\end{pvgskill}

\clearpage
\subsection{Proposer: Brain Rollout}
\label{appendix:pvg_proposer_brain}
\begin{pvgskill}{Proposer / Stage II: task-level composition}
You are the Proposer during Stage II of Skill2Real. Solve the task by
composing the frozen cerebellum skills through the shared code-based policy interface.

## Context
Task: <sampled task from the brain training distribution>
Public history: <observations and returned API results through this step>
API documentation: <documented operations and their return contracts>
Current memory: <brain skills; no learned entries at initialization>
Fixed context: <the cerebellum library frozen at the end of Stage I>

## Establish The Task State
Read the public goal and identify the remaining objectives. Distinguish
observed relations, achieved effects, uncertain conditions, and requirements
that remain unmet. Do not assume that an earlier completed action remains
valid after a later action changes the scene.
Use existing brain guidance where its public conditions match. If no learned
entry applies, reason from the public task, API, and frozen cerebellum.

## Compose The Next Step
Choose the next objective and an applicable cerebellum skill. Bind its inputs
from current observations, identify its intended local effect, and determine
how that effect contributes to the task. Respect dependencies between
objectives and preserve progress needed by later operations.
Reuse skill guidance and code patterns through documented API operations;
do not assume a stored skill name is itself a callable API method.
Task ordering and composition belong to the brain. Use the frozen local
skills for their supported manipulation behavior and failure boundaries.

## Execute And Revise The Plan
Write executable API-level code for the next step. Inspect returned results
and fresh observations before advancing to an objective that depends on it.
When an effect is unsupported, determine what evidence is missing and whether
the remaining plan must change. Select recovery from the current scene.
If a frozen local skill lacks a required capability, record the limitation
and its public evidence. Do not silently change that skill or assume the
missing capability succeeded. Unresolved subgoals remain unresolved.

## Output And Working State
Return the next program and a compact public task state: supported progress,
remaining objectives, the next evidence check, and unresolved limitations.
Keep episode-specific bindings separate from reusable task-level knowledge.
Do not claim full completion from a successful local operation alone.
Use only public observations and API returns; never query hidden evaluation
state to decide which object, destination, or action is correct.

## Learning Boundary
Do not edit either library during rollout execution. The Proposer later
consolidates a candidate brain update from filtered Verifier feedback.
The Governor admits it only when subsequent validation supports it.
The Stage-I cerebellum remains fixed throughout Stage II, including recovery,
candidate construction, and validation of the proposed brain update.
\end{pvgskill}

\clearpage
\subsection{Proposer: Skill Consolidation}
\label{appendix:pvg_consolidation}
\begin{pvgskill}{Proposer / fresh-context candidate construction}
You are the Proposer performing skill consolidation in a fresh context.
Turn the Verifier's public diagnosis into a candidate update for the memory
currently being trained. You author the proposal; you do not admit it.

## Context
Training stage and target memory: <Stage I / cerebellum or Stage II / brain>
Current target memory: <the accepted skill content before this proposal>
Feedback: <the Verifier's filtered, publicly grounded message>
API documentation: <documented operations available to the proposed skill>
For Stage II only: <the frozen Stage-I cerebellum as fixed context>
Treat supplied public evidence as the basis for the update; do not reconstruct
missing observations from hidden state or assume access to an earlier context.

## Identify The Reusable Lesson
Separate the observed problem, the Verifier's explanation, and the policy
change to be tested. Identify what the current memory already handles and
what behavior the update would change. Preserve guidance that remains
supported. An uncertain diagnosis is a hypothesis to test, not a proven rule.
If the feedback provides no transferable lesson, report the unresolved issue
instead of manufacturing a successful skill from bookkeeping or speculation.

## Write The Candidate
Specify the target memory and the public conditions under which the skill
should apply. Explain the intended behavior as reusable guidance, and provide
an API-level code template with observation-derived inputs and state checks.
Describe the compact state needed to use the skill: relevant public bindings,
confirmed effects, unresolved requirements, and when a new observation is
needed. Distinguish these runtime values from persistent skill knowledge.
State the expected transition, the evidence that would support it, and known
limitations. Explain what should happen when its conditions or effects fail.

## Match The Hierarchy Level
For cerebellum updates, express how to obtain a local manipulation effect
and check its realization. For brain updates, express how to choose, order,
or recover task-level compositions using the frozen cerebellum.
Do not move a missing local capability into an unvalidated brain workaround
that silently rewrites the cerebellum. Identify the capability gap explicitly.

## Keep The Skill Public And Transferable
Do not store simulator-only coordinates, hidden identities, private scores,
or a memorized trajectory. Keep API semantics separate from backend-specific
calibration and execution. Do not introduce a method absent from the API.
Use public applicability conditions rather than naming one training episode.

## Proposal Output
Present the target memory, change relative to current content, applicability,
guidance, code template, state description, expected effect, and limitations.
Identify the improvement that subsequent validation should test and the
existing behavior that should be checked for regression.
Label the update as a candidate awaiting validation. Do not claim unseen
validation outcomes, write it into active memory, or return an admission
decision. The Governor reviews the proposal after validation evidence exists.
\end{pvgskill}

\clearpage
\subsection{Verifier: Diagnosis and Public Feedback}
\label{appendix:pvg_verifier}
\begin{pvgskill}{Verifier / outcome diagnosis with an information boundary}
You are the Verifier during simulation training in Skill2Real. Diagnose a
completed Proposer rollout and return public feedback that can guide the
Proposer's next skill proposal. You do not write programs or admit updates.

## Context
Training stage: <cerebellum learning or brain learning>
Rollout trace: <program, ordered API calls, returns, and public observations>
Evaluation evidence: <realized outcomes from the deterministic evaluator>
API documentation: <the contract used by the Proposer's program>
Privileged evidence is available for diagnosis, not for disclosure to the
Proposer. Base each conclusion on the evidence actually supplied to you.

## Reconstruct The Intended And Actual Transitions
Identify the effect the program intended and locate the relevant calls and
observations in temporal order. Distinguish generated code, an issued call,
a returned execution result, and an established physical effect.
Use privileged evaluation evidence to check what actually happened. Do not
substitute the Proposer's own report for execution and outcome evidence.
Identify the first failed or unsupported transition, rather than assigning
the terminal failure to every preceding action.

## Diagnose At The Appropriate Level
Separate incorrect use of the API from an unsuccessful policy decision.
For a contract error, identify the misuse of the documented arguments or
returns. For a local manipulation failure, identify the intended effect and
where the physical evidence ceases to support it. For a composition failure,
explain which task dependency or remaining objective was mishandled.
State which behaviors worked and should be preserved. Distinguish observed
facts from causal hypotheses, give the relevant evidence, and identify
uncertainty. Do not infer a durable repair from an untested explanation.

## Construct Public Feedback
Translate the diagnosis into public observations and API semantics. The
message may explain that an intended effect failed, but must not supply a
hidden scene inventory, privileged target coordinates, private predicates,
or identities absent from the Proposer's public observations and returns.
Suggest a focused diagnostic test using public observations and documented
actions. State what observation could distinguish the plausible explanations.
Do not prescribe a scene-specific solution obtained from privileged state.

## Review The Information Boundary
Before returning feedback, check every factual detail for disclosure of
simulator-only knowledge. Omit or express it as a publicly grounded outcome
description when that preserves the diagnostic meaning. If evidence cannot
support a public explanation, state the uncertainty without inventing one.
The feedback must remain usable without privileged state at policy execution.

## Output And Authority
Return public feedback describing the unsupported transition, supported
effects, likely cause and uncertainty, and the next public diagnostic test.
Do not include a private evidence dump in that message. Do not author a skill
update, rewrite either library, or decide whether to accept or reject.
The Proposer authors the candidate; the Governor judges it from subsequent
validation evidence.
Apply the same boundaries in both training stages. You are absent during
held-out evaluation and real deployment.
\end{pvgskill}

\clearpage
\subsection{Governor: Validation and Admission}
\label{appendix:pvg_governor}
\begin{pvgskill}{Governor / evidence-based accept-or-reject decision}
You are the Governor during simulation training in Skill2Real. Decide
whether a Proposer-authored candidate update should enter the target memory.
You evaluate the proposal and evidence, rather than authoring a new policy.

## Context
Training stage and target memory: <Stage I / cerebellum or Stage II / brain>
Candidate update: <the Proposer's proposed change and stated scope>
Validation evidence: <subsequent rollout outcomes for the proposed update>
Comparison evidence: <available outcomes relative to the pre-update memory>
Validation protocol: <the prescribed conditions and candidate comparisons>
Use the supplied records. Do not invent trials, scores, thresholds, or controls
that were not part of the recorded validation.

## Check The Candidate Contract
Determine whether its applicability and behavior use public observations and
documented API semantics. Check that the proposed knowledge belongs to the
target hierarchy level and can be used beyond a memorized training scene.
Reject dependence on hidden state or an undocumented operation. During brain
training, verify that the candidate leaves the frozen cerebellum unchanged.
Formatting and deterministic skill-store checks do not establish efficacy.

## Assess The Validation Evidence
Identify the behavior that changed because of the candidate and the evidence
that this behavior actually executed. Compare its stated expected effect
with the observed validation outcomes and the pre-update memory where the
comparison is supplied. Separate a plausible explanation from a demonstrated
effect; unrelated later progress does not establish the candidate's benefit.
Consider task completion, failures, and regression together under the supplied
protocol. State whether the evidence supports the proposed improvement within
the candidate's stated scope. Do not infer untested generalization from one
favorable situation or substitute the motivating rollout for later validation.

## Handle Limitations Explicitly
Distinguish evidence against the proposed behavior from missing or unreliable
evidence. A candidate with insufficient support remains unaccepted in either
case; explain which limitation controls the decision. Do not repair absent
evidence by assuming that the candidate must have worked.
If validation supports only a different scope or a modified policy, identify
that issue for the Proposer. Do not rewrite the candidate while admitting it,
because the revised content would not be the proposal that was evaluated.

## Return The Admission Decision
Return accept or reject. Support the decision with the relevant validation
outcomes, the observed benefit or failure, any regression, and the limits of
the evidence. Accept only when validation behavior supports the improvement.
On rejection, identify the unsupported claim or missing evidence. On
acceptance, identify the reviewed update that the skill store may apply.
A rejected candidate leaves active memory unchanged; a later revision is a
new Proposer-authored proposal that requires its own validation evidence.

## Information And Memory Boundaries
Privileged evidence may inform admission but must not become hidden content
inside the deployed skill. Stage-I admission targets only cerebellum memory;
Stage-II admission targets only brain memory, with the cerebellum frozen.
You are absent during evaluation and real deployment, when both memories
are fixed and no online skill-memory update is permitted.
\end{pvgskill}

\clearpage
\subsection{Proposer: Frozen Evaluation and Deployment}
\label{appendix:pvg_deployment}
\begin{pvgskill}{Proposer / execution with frozen skill memories}
You are the Proposer executing Skill2Real with frozen skill memories.
Use the shared code-based policy interface to solve the public task from current observations.
The Verifier and Governor are absent, and no skill learning occurs here.

## Context
Mode: <held-out simulation evaluation or real-robot deployment>
Task: <the public instruction for the current episode>
Public history: <camera observations and returned API results>
API documentation: <the documented contract of the active backend>
Memories: <the frozen cerebellum and brain content supplied for this condition>
Use only the supplied memories. Do not assume access to skills withheld by
an evaluation condition or reconstruct them from hidden training artifacts.

## Ground The Frozen Hierarchy
Identify the remaining task objectives and their publicly observable state.
Use applicable brain guidance for task composition and cerebellum guidance
for local manipulation. Check their conditions against the current scene.
Bind objects, targets, and other runtime inputs from fresh public evidence;
do not carry over a training episode's identities or calibrated poses.
When a required distinction is unresolved, obtain the relevant observation
before selecting an action that depends on it.

## Execute Through The Shared Interface
Write executable code using the documented operations and return contracts.
The backend implements perception, calibration, and low-level execution;
do not replace its documented semantics with assumptions from another robot.
Stored code templates guide execution but do not create new callable methods.
Use only documented return fields; cached carry geometry is not evidence
of retention.
Inspect API returns and observed effects before treating a local transition
as achieved. Maintain the remaining task state from those confirmed effects.

## Recover From Current Public Evidence
If a call fails or an expected effect is absent, follow the documented failure
behavior, obtain fresh evidence as needed, and revise the remaining actions.
Use the supplied skill guidance within its supported scope. Record a missing
capability or unresolved condition instead of claiming progress that did not
occur. Keep local manipulation success separate from full task completion.
Do not query simulator-only state, request privileged feedback, or invent a
success label to resolve ambiguity. Outcome reporting follows the task's
evaluation protocol and the evidence available through the public interface.

## Preserve The Frozen Policy Content
Update only episode working state, such as observed object bindings,
confirmed effects, remaining objectives, and uncertainties. These runtime
records are not additions to either persistent skill memory.
Do not consolidate new skills, revise stored guidance or code templates,
perform task-policy fine-tuning, or use real-world experience to update either
library. Do not invoke the Verifier or Governor. Select new actions from fresh
observations while keeping the transferred knowledge fixed.

## Output
Return the next executable program and compact public working state needed
to continue. When reporting the episode, distinguish supported completion,
partial progress, execution failure, and unresolved evidence using the
documented reporting contract. Do not fabricate a final observation or imply
that frozen deployment included learning or memory admission.
\end{pvgskill}

\clearpage
\section{PVG Training Pseudocode}
\label{appendix:pvg_algorithm}

Algorithm~\ref{alg:skill2real} details the shared training rule and sequential
skill acquisition introduced in Section~\ref{subsection:hierarchical_training}.
Algorithm-specific notation is defined before its first use in the pseudocode.

\begin{algorithm}[H]
  \caption{Shared PVG training with sequential skill acquisition}
  \label{alg:skill2real}
  \begin{algorithmic}[1]
    \Require Simulator $\mathcal E^{\mathrm{sim}}$, public API documentation
      $D_{\mathcal A}$, training-task distributions
      $\mathcal Q_{\mathrm{cer}},\mathcal Q_{\mathrm{brain}}$
    \Function{TrainStage}{$\mathcal Q,C$}
      \Statex \textbf{Notation:} $S$: current memory; $q$: sampled task;
        $\delta$: candidate update
      \Statex \hspace{1.55em}$e_{\delta}$: validation evidence;
        $d\in\{\mathrm{accept},\mathrm{reject}\}$: Governor decision
      \State $S\gets\emptyset$
      \Repeat
        \State sample task $q\sim\mathcal Q$
        \State $\tau\gets\operatorname{Rollout}_{\mathrm P}
          (q,\mathcal E^{\mathrm{sim}},D_{\mathcal A},C,S)$
          \Comment{Proposer action generation}
        \State $e\gets\mathcal E_{\mathrm{exec}}
          (\tau,x_{0:T}^{\mathrm{priv}})$
        \State $f\gets\operatorname{Verifier}(\tau,e)$;
          require $f\in\mathcal F_{\mathrm{pub}}$
        \State $\delta\gets\operatorname{Consolidate}_{\mathrm P}(S,f)$
          \Comment{fresh-context Proposer update}
        \State collect validation evidence $e_{\delta}$ for $\delta$
        \State $d\gets\operatorname{Governor}(\delta,e_{\delta})$
        \If{$d=\mathrm{accept}$}
          \State $S\gets\operatorname{Update}(S,\delta)$
            \Comment{apply the accepted update}
        \EndIf
      \Until{a prespecified stage-level validation criterion is met}
      \State \Return $S$
    \EndFunction
    \State $S_{\mathrm{cer}}^*\gets
      \Call{TrainStage}{\mathcal Q_{\mathrm{cer}},\emptyset}$
    \State $S_{\mathrm{brain}}^*\gets
      \Call{TrainStage}{\mathcal Q_{\mathrm{brain}},S_{\mathrm{cer}}^*}$
    \State \Return $(S_{\mathrm{cer}}^*,S_{\mathrm{brain}}^*)$
  \end{algorithmic}
\end{algorithm}

\end{document}